\pdfoutput=1
\documentclass[sigconf,nonacm]{acmart}

\AtBeginDocument{%
  }

\usepackage{algorithm}
\usepackage{algorithmic}

\begin{document}

\title{From Pixels to Temporal Correlations: Learning Informative Representations for Reinforcement Learning Pre-training}

\author{Jinwen Wang}
\orcid{0009-0004-6280-0801}
\affiliation{%
  \department{School of Computer Science \& Technology,}
  \institution{Beijing Jiaotong University}
  \city{Beijing}
  \country{China}
}
\email{jw.wang@bjtu.edu.cn}

\author{Youfang Lin}
\orcid{0000-0002-5143-3645}
\affiliation{%
  \department{Beijing Key Laboratory of Traffic Data Mining and Embodied Intelligence,}
  \institution{Beijing Jiaotong University}
  \city{Beijing}
  \country{China}
}
\email{yflin@bjtu.edu.cn}

\author{Xiaobo Hu}
\orcid{0000-0001-6541-2784}
\affiliation{%
  \institution{Beijing Jiaotong University}
  \city{Beijing}
  \country{China}
}
\email{xiaobohu@bjtu.edu.cn}

\author{Siyu Yang}
\orcid{0009-0008-4230-7656}
\affiliation{%
  \institution{Beijing Jiaotong University}
  \city{Beijing}
  \country{China}
}
\email{yangsiyu@bjtu.edu.cn}

\author{Sheng Han}
\orcid{0000-0003-4049-3690}
\affiliation{%
  \institution{Beijing Jiaotong University}
  \city{Beijing}
  \country{China}
}
\email{shhan@bjtu.edu.cn}

\author{Shuo Wang}
\authornote{Corresponding author.}
\orcid{0000-0001-6599-3638}
\affiliation{%
  \institution{Beijing Jiaotong University}
  \city{Beijing}
  \country{China}
}
\email{shuo.wang@bjtu.edu.cn}

\author{Kai Lv}
\orcid{0000-0001-6533-5176}
\affiliation{%
  \department{Beijing Key Laboratory of Traffic Data Mining and Embodied Intelligence,}
  \institution{Beijing Jiaotong University}
  \city{Beijing}
  \country{China}
}
\email{lvkai@bjtu.edu.cn}

 \renewcommand{\shortauthors}{Jinwen Wang et al.}

\begin{abstract} 
Unsupervised pre-training on large-scale datasets has demonstrated significant potential for improving the sample efficiency and performance of Reinforcement Learning (RL). 
Given the large-scale action-free internet videos, existing methods utilize single-step transition prediction and image reconstruction to learn representations.
However, these methods prefer to preserve large-proportion stationary information in the pixel space, neglecting small but crucial information. 
To preserve enough information in the representation, it is essential to pay equal attention to each element in videos.
Specifically, we propose a temporal correlation space to distinguish each element. 
For implementation, we introduce the Multi-scale Temporal Contrastive Learning (MTCL) method to model multi-scale temporal correlations separately. 
This approach can balance the attention of different elements and yield more informative representations, effectively supporting policy learning in various downstream tasks. 
Experimental results demonstrate that our method improves sample efficiency and asymptotic performance across various downstream tasks.
\end{abstract}

\ccsdesc[500]{Computing methodologies~Computer vision tasks}

\keywords{Reinforcement Learning Pre-training; Temporal Correlation Space; Multi-scale Modeling; Informative Representations}


\maketitle

\section{Introduction}
Deep Reinforcement Learning (RL) has achieved remarkable success in various fields~\cite{DBLP:journals/jmlr/LevineFDA16,DBLP:conf/nips/ZhouLJL23,DBLP:conf/corl/HuangWZL0023,DBLP:journals/tmm/WangWHLL23, DBLP:conf/aaai/WangWHWLL24}. 
However, RL often starts from scratch, requiring extensive interaction experience to learn effective policies.
Inspired by the success of the \textit{pre-training and fine-tuning} paradigm in Computer Vision (CV)~\cite{DBLP:conf/iclr/DosovitskiyB0WZ21,DBLP:conf/cvpr/He0WXG20,DBLP:conf/cvpr/HeCXLDG22} and Natural Language Processing (NLP)~\cite{DBLP:conf/naacl/DevlinCLT19,DBLP:conf/nips/YangDYCSL19,DBLP:conf/nips/BrownMRSKDNSSAA20}, exploring pre-training for RL with large-scale internet video data is highly appealing for improving sample efficiency and performance~\cite{DBLP:conf/corl/NairRKF022,DBLP:conf/corl/RadosavovicXJAM22,DBLP:conf/nips/YuanXYWWGX22}.

\begin{figure}[]
  \centering
  \includegraphics[width=0.85\linewidth]{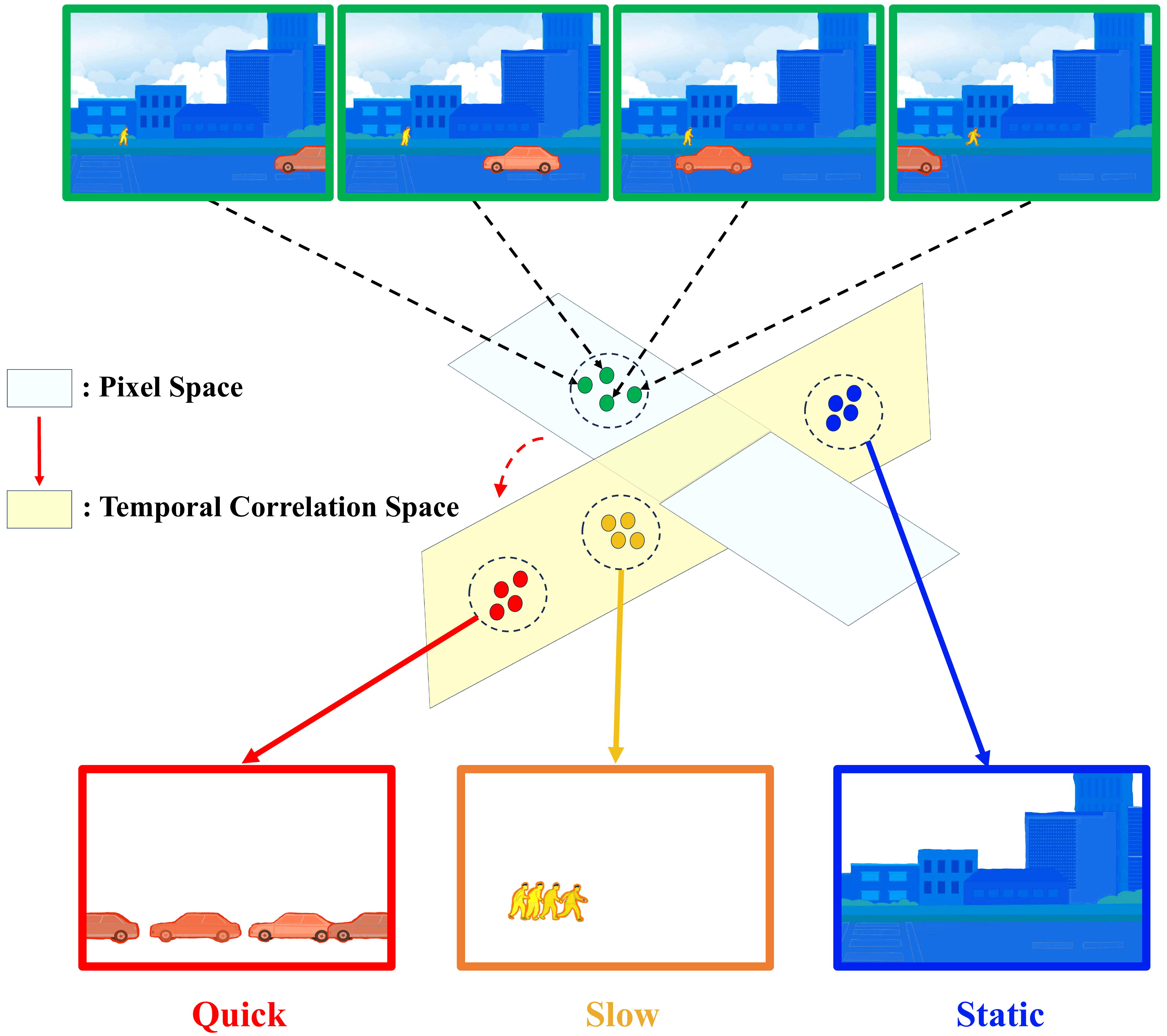}
  \caption{\textbf{Illustration of our method.} 
  We convert videos from the pixel space to the temporal correlation space where elements are inherently separable. By equally attending to different elements, we learn more informative representations that effectively support various downstream tasks.
  }
  \label{fig. intro}
\end{figure}

Since large-scale internet videos lack action labels, recent methods~\cite{DBLP:conf/icml/SeoLJA22,DBLP:conf/nips/0001MDL23,DBLP:conf/eccv/ZhangKSC24} have explored using single-step transition prediction and image reconstruction to learn representations in unsupervised pre-training.
However, these methods tend to focus on elements with large-proportion pixels~\cite{DBLP:conf/icra/OkadaT21}, and the transition prediction may fail to trivial solutions~\cite{DBLP:conf/iclr/HansenJSAAEPW21}. 
As a result, these representations omit elements with a small pixel proportion, which may be crucial for decision-making, especially when prior knowledge of downstream tasks is lacking.
Therefore, it is important to preserve more elements in the representation. 

As mentioned above, elements with varying pixel proportions contribute differently to the representation. 
Therefore, improving attention to critical but small elements is essential. 
However, distinguishing different elements \textit{in the pixel space} is challenging.
Here comes the question: \textit{can we construct a new space to reorganize the information for distinguishing elements? }
Inspired by~\cite{DBLP:journals/corr/abs-2403-13765}, temporal correlation, related to the motion velocity, is a new perspective for distinguishing different elements.  
In this paper, we convert information from the pixel space to the temporal correlation space, providing a foundation for learning more informative representations. 

As shown in Figure~\ref{fig. intro}, elements in the video include vehicles, pedestrians, and buildings. 
Since static elements like buildings occupy the majority of the pixels, different frames are close to each other \textit{in the pixel space}, resulting in the challenge of distinguishing each element.
In contrast, different elements have varying motion velocities, making them inherently separable \textit{in the temporal correlation space}.

In this paper, we learn representations for RL pre-training \textit{from pixels to temporal 
correlations}. 
Based on this new space, we propose Multi-scale Temporal Contrastive Learning (MTCL), a method that separately models multi-scale temporal correlations to ensure equal attention to different elements in the video.
Specifically, we assign a distinct contrastive learning objective to each temporal correlation scale, ensuring balanced attention across various elements.
Due to the differences in selecting positive and negative samples, we divide these contrastive learning objectives into two parts: one captures multi-scale motion-aware information, while the other focuses on static appearance-aware information, such as background, color, and texture.
Our method addresses the issue of information omission in representations, resulting in more informative representations that effectively support policy learning across various downstream tasks.

We conduct extensive experiments on three different downstream benchmarks: DMControl Remastered, Meta-World, and CARLA. 
The results show that our method (MTCL) significantly improves sample efficiency and asymptotic performance.

Our main contributions are summarized as follows:
\begin{itemize}
\item We propose a temporal correlation space where elements are inherently separable and independently model multi-scale temporal correlations to learn more informative representations.

\item In practice, we establish a series of contrastive learning objectives across different temporal correlation scales, composed of Multi-scale Motion-aware Learning (MML) and Static Appearance-aware Learning (SAL).

\item Extensive experimental results on three benchmark tasks demonstrate that our method significantly improves sample efficiency and asymptotic performance in downstream tasks, achieving state-of-the-art results.
\end{itemize}

\section{Related Work}
\subsection{Unsupervised Pre-training for RL}
Recent researches~\cite{DBLP:conf/icml/GhoshBL23,DBLP:conf/corl/RadosavovicXJAM22,DBLP:conf/iclr/YeZAG23} have shown that unsupervised pre-training for RL significantly improves sample efficiency and asymptotic performance across various downstream decision-making tasks.

APV~\cite{DBLP:conf/icml/SeoLJA22} adopts a two-stage learning process of \textit{pre-training and fine-tuning}. 
First, it pre-trains an action-free world model with videos from RLBench~\cite{DBLP:journals/ral/JamesMAD20}. 
During fine-tuning, an action-conditional dynamics model is stacked on the pre-trained model to learn policies. 
This approach improves sample efficiency and performance across various downstream tasks. 
Since the dataset utilized during the pre-training needs to be sampled from a specific domain, APV struggles to leverage large-scale and diverse data.
Similarly, IPV~\cite{DBLP:conf/nips/0001MDL23} adopts the \textit{pre-training and fine-tuning} paradigm and further introduces rich internet videos for pre-training. 
It proposes a Contextualized World Model to address the complexity of internet videos.
PreLAR~\cite{DBLP:conf/eccv/ZhangKSC24} introduces a learnable action representation to utilize action-free videos for pre-training world models.

However, the above methods adopt similar architecture for learning representation, \textit{i.e.}, one-step transition prediction and image reconstruction.
These pixel-based methods tend to preserve large-proportion stationary information, neglecting small but crucial information in representations.
To address this issue, we propose converting videos from the pixel space to the temporal correlation space and then independently modeling multi-scale temporal correlations to ensure equal attention to different elements.

\subsection{Model-Based Reinforcement Learning}
Model-Based Reinforcement Learning (MBRL) improves sample efficiency by building world models of the environment to generate hypothetical trajectories. 
The high sample efficiency of MBRL algorithms highlights their significant potential in tackling sequential decision-making problems in complex scenarios~\cite{DBLP:conf/iclr/HafnerL0B21,DBLP:conf/nips/YeLKAG21,DBLP:journals/corr/abs-2202-09481,DBLP:journals/corr/abs-2502-01591}.

PlaNet~\cite{DBLP:conf/icml/HafnerLFVHLD19} introduces a Recurrent State Space Model (RSSM) to learn environment dynamics from images and selects actions through rapid online planning in the latent space.
Dreamer~\cite{DBLP:conf/iclr/HafnerLB020} proposes a latent dynamics model using a Variational Autoencoder (VAE), which encodes observations and actions into compact latent states and then effectively learns policies from imagined latent trajectories.
DreamerV2~\cite{DBLP:conf/iclr/HafnerL0B21} is the first RL algorithm to achieve human-level performance on the Atari benchmark solely by learning behaviors within a separately trained world model.
TD-MPC~\cite{DBLP:conf/icml/HansenSW22} introduces local trajectory optimization and long-term return estimation to handle the high cost and accuracy issues of long-horizon planning.
DreamerV3~\cite{DBLP:journals/corr/abs-2301-04104}, introduces a range of robustness techniques based on normalization, balancing, and transformations, outperforming specialized approaches across more than 150 diverse tasks.

Despite these advancements, traditional MBRL approaches typically start learning from scratch, making it difficult to quickly adapt to different downstream tasks. 
In contrast, our method pre-trains a reusable representation that can be broadly applied across various downstream tasks, improving sample efficiency and performance.

\section{Preliminaries}
\subsection{Problem formulation}
In visual reinforcement learning, due to the partial observability of images, the interaction between the agent and the environment is modeled as a Partially Observable Markov Decision Process (POMDP), represented by the tuple \( \mathcal{M} = \langle \mathcal{S}, \mathcal{O}, \mathcal{A}, \mathcal{P}, \mathcal{R}, \gamma \rangle \), where \( \mathcal{S} \) is the state space, \( \mathcal{O} \) is the observation space, \( \mathcal{A} \) is the action space, \( \mathcal{P}: \mathcal{S} \times \mathcal{A} \mapsto \mathcal{S} \) is the state transition function, \( \mathcal{R}: \mathcal{S} \times \mathcal{A} \mapsto  \mathbb{R} \) is the reward function, \( \gamma \in [0,1) \) is the discount factor.
The objective is to learn the optimal policy \( \pi^* = \arg\max_{\pi} \mathbb{E}_{a_t \sim \pi, s_t \sim \mathcal{P}} \left[ \sum_{t=0}^{T-1} \gamma^t \mathcal{R}(s_t, a_t) \right] \), starting from the initial state \( s_0 \in \mathcal{S} \) and taking actions \( a_t \) chosen by the policy \( \pi_{\theta} (\cdot \mid s_t) \),  parameterized by \( \theta \).
Here, \( T \) is the horizon of the trajectory.

\subsection{Latent Dynamics Models}
Dreamer~\cite{DBLP:conf/iclr/HafnerLB020} introduces a latent dynamics model composed of four main components:
\begin{equation}
\begin{aligned}
  &\text{Representation model:} &  \quad &q_\theta(z_t | z_{t-1}, a_{t-1}, o_t) \\
  &\text{Transition model:} &  \quad &p_\theta(\hat{z}_t | z_{t-1}, a_{t-1}) \\
  &\text{Reward model:} & \quad &p_\theta(r_t | z_t) \\
  &\text{Image decoder:} & \quad &p_\theta(o_t | z_t),
\end{aligned}
\end{equation}
The representation model encodes observations $o_t$ and actions $a_{t-1}$ into compact latent representations $z_t$ with Markovian transitions.
The image decoder ensures that the encoded representations $z_t$ retain as much observation information as possible.
The transition model effectively predicts future representation $\hat{z}_t$ by approximating the representation model, and the reward model predicts the reward $r_t$ for a given representation $z_t$.

\subsection{Contextualized World Models}

The Contextualized World Model (ContextWM)~\cite{DBLP:conf/nips/0001MDL23} provides a pathway for pre-training with rich internet videos by separately modeling the context, effectively addressing the high complexity of internet videos. 
Compared to the latent dynamic model, ContextWM introduces two main structural improvements.

Firstly, the image decoder \( p_\theta (o_t | z_t, c) \) generates observations \( o_t \) based not only on the current representation \( z_t \) but also on context variables \( c \). 
The decoder is further enhanced by incorporating cross-attention mechanisms~\cite{DBLP:conf/nips/VaswaniSPUJGKP17} and residual-connection~\cite{DBLP:conf/cvpr/HeZRS16}.

Secondly, ContextWM introduces a dual reward predictor to address the issue where traditional video-based intrinsic rewards \( r_t^{\text{int}} \)~\cite{DBLP:conf/icml/SeoLJA22} can distort the regression of pure rewards \(r_t\) during fine-tuning.  
Specifically, the dual reward predictor includes a behavior reward predictor \( p_\phi (r_t + \lambda r_t^{\text{int}} | s_t) \) and a representative reward predictor \( p_\varphi (r_t | s_t) \).

The overall optimization objective of the ContextWM is:
\begin{equation}
\begin{aligned}
\mathcal{L}_{\text{CWM}}  = \mathbb{E}_{q_\phi, q_\theta} \Big[ \sum_{t=1}^T \Big( - \ln p_\theta (o_t | s_t, c)  - \beta_r \ln p_\varphi (r_t | s_t)  \\
- \ln p_\phi (r_t + \lambda r_t^{\text{int}} | s_t)+ \beta_z \mathcal{L}_z + \beta_s \mathcal{L}_s \Big) \Big],
\end{aligned}
\label{eq.2}
\end{equation}
where \(\mathcal{L}_z\) represents the action-free KL loss, and \(\mathcal{L}_s\) represents the action-conditional KL loss, expressed as:
\begin{equation}
\begin{aligned}
&\mathcal{L}_z = \mathrm{KL} \left[ q_\theta(z_t | z_{t-1}, o_t) \parallel p_\theta(\hat{z}_t | z_{t-1}) \right],\\
&\mathcal{L}_s = \mathrm{KL} \left[ q_\phi(s_t | s_{t-1},a_{t-1}, z_t) \parallel p_\phi(\hat{s}_t | s_{t-1}, a_{t-1}) \right].
\end{aligned}
\end{equation}

\section{Method}
In this section, we introduce the temporal correlation space to identify different elements in the video and then independently model multi-scale temporal correlations by Multi-scale Temporal Contrastive Learning (MTCL). 
In practice, we set a series of contrastive learning objectives which can be categorized into two types: Multi-scale Motion-aware Learning (MML) and Static Appearance-aware Learning (SAL).
Additionally, we demonstrate how to integrate the MML and SAL into the framework.

\begin{figure*}
  \centering
  \includegraphics[width=\linewidth]{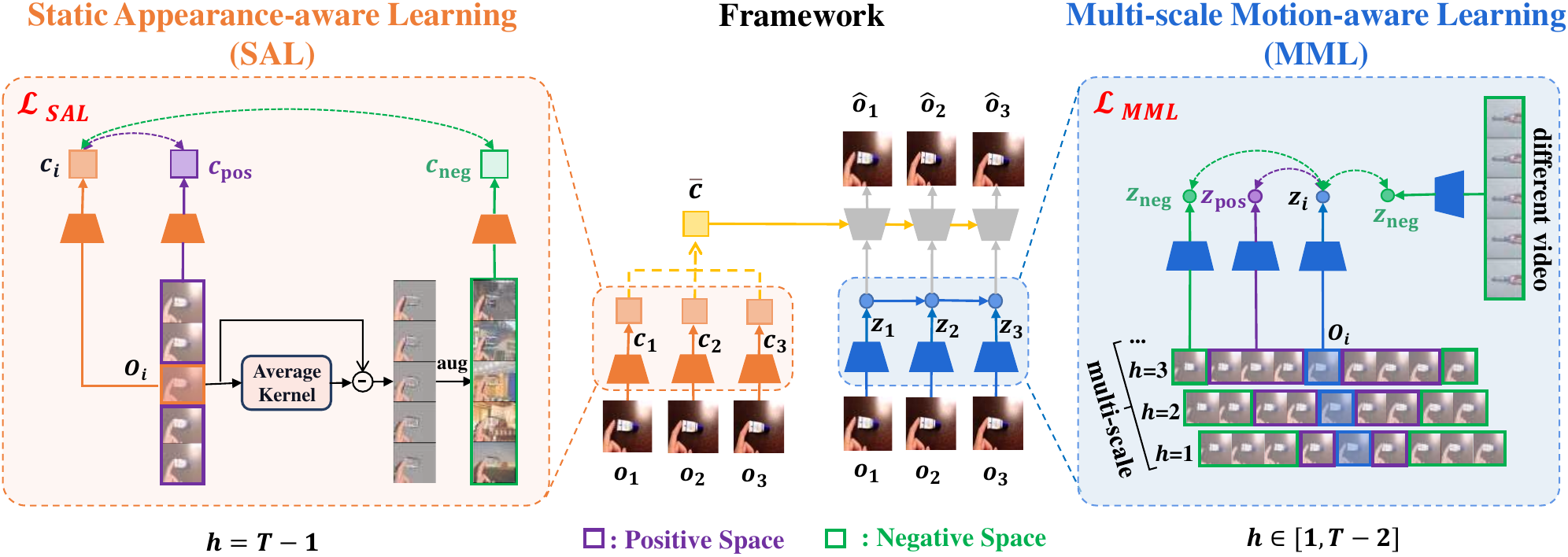}
  \caption{\textbf{Overview of our model.} Building on an action-free world model framework, we independently model multi-scale temporal correlations in videos. 
  Our method is composed of Multi-scale Motion-aware Learning (MML) and Static Appearance-aware Learning (SAL).
  The MML objective is applied to the observation encoder and the SAL objective is applied to the context encoder. Average pooling is applied to the context variable \(c\) over the sequence dimension to enhance the decoder.}
  \label{fig:Overview}
\end{figure*}

\subsection{Multi-scale Temporal Correlation Modeling}
To learn more informative representations, it is crucial to pay equal attention to each element in the video.
The video can be regarded as a mixture distribution of different elements.
An intuitive approach is maximizing the Evidence Lower Bound (ELBO) by variational inference~\cite{DBLP:journals/corr/BleiKM16} to approximate the mixture distribution.
However, it is difficult to distinguish each element in the pixel space.

In contrast, we propose a temporal correlation space where elements are inherently separable by different motion velocities.
Furthermore, we prove that there exists a connection between the ELBO of variational inference and contrastive learning.
The detailed proof is exhibited in Appendix 3.
Therefore, we utilize contrastive learning to independently model multi-scale temporal correlations, ensuring balanced attention to each element.

In practice, we establish a series of contrastive learning objectives. 
Each objective corresponds to a specific temporal correlation scale and can be formalized as follows:

\begin{equation}
\mathcal{L}_{h} = -\mathbb{E} \left[ \log \frac{e^{d(z_i, z_j)}}{e^{d(z_i, z_j)} + e^{d(z_i, z_k)}} \right],
\end{equation}
where \( h \) ranges from 1 to \(T-1\), representing different temporal correlation scales.
\( z_i \) is the representation of current frame \( i \), \( z_j \) comes from the positive sample space \( \delta^{+}_{i} \), and \( z_k \) comes from the negative sample space \( \delta^{-}_{i} \).
The selection of the positive and negative sample space is related to the scale \( h \).
As temporal correlation increases (i.e., as \( h \) becomes smaller), the range of the positive sample space \( \delta^{+}_{i} \) becomes narrower, meaning that the corresponding element changes within shorter time intervals.
The \( d(\cdot) \) denotes the measure of similarity, which in this paper is defined as the negative L2 distance.
The overall contrastive learning objective is obtained by summing these individual objectives:
\begin{equation}
\mathcal{L}_{\text{total}} = \sum_{h=1}^{T-1} \mathcal{L}_{h}.
\end{equation}

Minimizing \(\mathcal{L}_{\text{total}}\) effectively captures temporal correlations at various scales within the sequence. 
Due to the differences in selecting positive and negative sample spaces, we divide the overall optimization objective into two parts: Multi-scale Motion-aware Learning (MML) and Static Appearance-aware Learning (SAL).

\subsection{Multi-scale Motion-aware Learning}
Multi-scale Motion-aware Learning (MML) is illustrated on the right of Figure~\ref{fig:Overview}.
We utilize \(\mathcal{L}_{1} \sim \mathcal{L}_{T-2}\) to capture motion elements at different scales within a video.
Given a video of \( T \) frames, the positive sample space \( \delta^{+}_{i} \) is defined as the frames within a horizon of \( h \) from the current frame \( i \). 
The negative sample space \( \delta^{-}_{i} \) consists of two parts: frames that are beyond the \( h \) horizon from frame \( i \) within the same video, and frames from different videos.
This is because the temporal correlation between different trajectories is generally lower than those within the same trajectory.
So \(\mathcal{L}_{h}\) captures the specific element that changes at intervals of \( h \) frames.


In practice, we randomly sample a temporal correlation scale \( h  \in [1, T-2] \), which exhibits a specific level of temporal correlation.
Positive and negative samples are randomly sampled from their respective spaces according to the selected scale, and the final optimization objective for our MML can be formalized as follows:
\begin{equation}
\mathcal{L}_{\text{MML}} = -\sum\limits_{b \in B} \log \frac{e^{d(z_i^b, z_j^b)}}{e^{d(z_i^b, z_j^b)} + e^{d(z_i^b, z_k^b)} + e^{d(z_i^b, z_i^{\neq b})}} ,
\label{eq.6}
\end{equation}
where \( z_j^b \) represents samples randomly selected from the positive sample space, \( z_k^b \) are samples beyond the positive sample space within the same video, and \( z_i^{\neq b} \) denotes samples from different videos. 
Notably, \( z_k^b \) and \( z_i^{\neq b} \) both represent the negative samples.

MML effectively captures elements in videos with different motion velocities. 
This approach addresses the limitations of previous methods, which may overlook motion elements with a small pixel proportion.
Therefore, it provides a solid foundation for policy learning in various downstream tasks.

\begin{figure*}
  \centering
  \includegraphics[width=\linewidth]{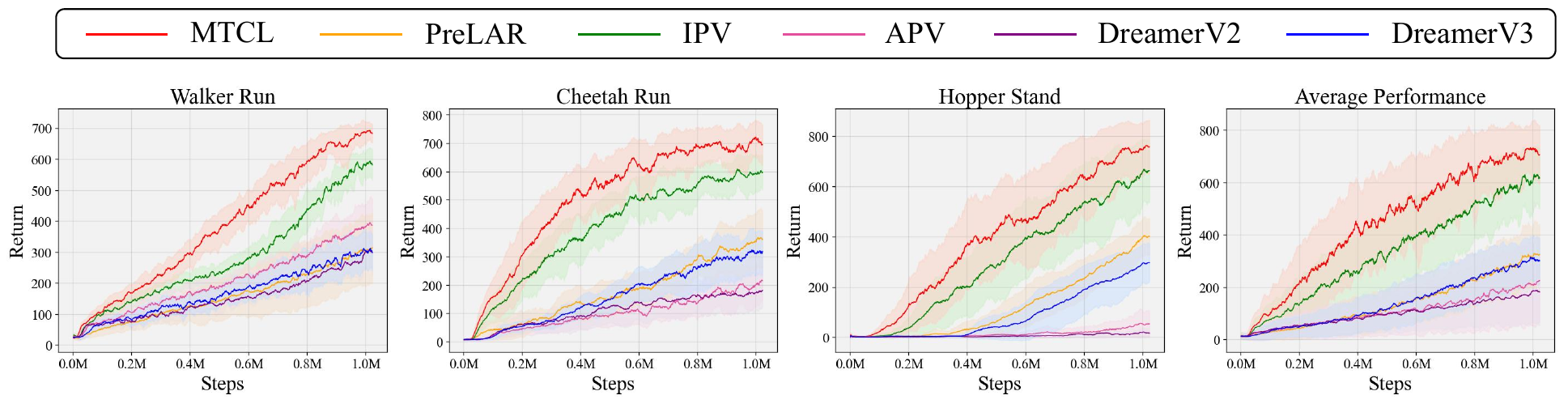}
  \caption{\textbf{Learning curves on DMControl Remastered. } We present the learning curves of our method (MTCL) compared to baselines across three tasks. Additionally, we report the average performance of each algorithm across the three tasks.}
  \label{result of DMControl Remastered}
\end{figure*}

\subsection{Static Appearance-aware Learning}
Static Appearance-aware Learning (SAL) captures static appearance-aware features in videos, such as background, color, and texture.
The architecture of SAL is demonstrated on the left of Figure~\ref{fig:Overview}.
Since static elements remain consistent throughout the entire video, the entire current video falls within the positive sample space \( \delta^{+}_{i} \), leaving no negative samples when optimizing \(\mathcal{L}_{T-1}\).


Moreover, it is not suitable for utilizing frames from different videos as negative samples.
In downstream tasks, different trajectories typically belong to the same domain and share similar static appearance features.
As a result, there are no suitable frames across all trajectories that can serve as effective negative samples for \(\mathcal{L}_{T-1}\).

To address this issue, we introduce a data augmentation technique that maximally disturbs the static pixels in the video, generating augmented images to form the negative sample space \( \delta^{-}_{i} \).
Specifically, given a batch of video sequences, we randomly select two different frames \([o_i, o_j]^{(1:B)}\) from each sequence and apply data augmentation to generate augmented images \([o_i',o_j']^{(1:B)}\). 
We then optimize the following loss objective:
\begin{equation}
\mathcal{L_{\text{SAL}}} = -\sum\limits_{b \in B} \log \frac{e^{d(c_i^b, c_j^b)}}{e^{d(c_i^b, c_j^b)} + e^{d(c_i^b, c_i^{b'})} + e^{d(c_j^b, c_j^{b'})}},
\label{eq.7}
\end{equation}
where \( c_i^b \) and \( c_j^b \) are the context variables extracted from the original frames,  with \( c_j^b \) serving as the positive sample for \( c_i^b \).
The  \( c_i^{b'} \) and \( c_j^{b'} \) are context variables extracted from the augmented images, serving as the negative samples for \( c_i^b \) and \( c_j^b \), respectively.


For data augmentation, we first calculate the temporal change rate for each pixel to determine the key pixel mask \( M_\rho \). 
We then use images \(\tilde{o}_i\) sampled from the Places dataset~\cite{DBLP:journals/pami/ZhouLKO018} to maximally perturb the non-key static pixels. 
The Places is widely used for random overlays~\cite{DBLP:conf/nips/HansenSW21}.
Thus, the data augmentation process can be represented as:
\begin{equation}
\text{aug}(o_i) = M_\rho \odot o_i + (1 - M_\rho) \odot \tilde{o}_i,
\end{equation}
where \(\odot\) denotes the Hadamard product.

By modeling the static temporal correlation, SAL effectively captures appearance-aware features in videos.
This approach reduces the complexity of internet videos by filtering out abundant static elements, allowing MML to focus more on moving elements with various motion velocities.
As a result, our method effectively learns more informative representations, which better supporting policy learning for various downstream tasks.

\subsection{Integration of MML and SAL}
In this paper, similar to APV~\cite{DBLP:conf/icml/SeoLJA22}, IPV~\cite{DBLP:conf/nips/0001MDL23} and PreLAR~\cite{DBLP:conf/eccv/ZhangKSC24}, we adopt the widely-used \textit{pre-training and fine-tuning} paradigm: first, we pre-train an action-free world model using internet video data, and then, during the fine-tuning stage, we stack an action-conditional world model on top of the pre-trained model to learn the policy.

As shown in Figure~\ref{fig:Overview}, our method independently models temporal correlations at different scales.
Due to the differences in selecting positive and negative sample spaces, the overall optimization objective is divided into two parts.
The MML objective is applied to the observation encoder to extract the latent representation \(z\) from each video frame.
The SAL objective is applied to the context encoder, which helps extract static contextual information \(c\) from complex scenes.


Next, we perform average pooling over the sequence dimension on the obtained static context variables \(c\) to derive the overall static context representation \(\bar{c}\):
\begin{equation}
\bar{c} = \text{mean}(c_{1:T}).
\end{equation}
Finally, we use the latent representation \(z_t\) with the overall static context representation \(\bar{c}\) for reconstruction.
\begin{algorithm}[tb]
	\caption{Multi-scale Temporal Correlation Pre-training} 
	\label{alg1} 
	\begin{algorithmic}[1]
		\STATE $\theta$: Initialize parameters of action-free dynamics model, image encoder, and decoder randomly
        \STATE Load internet video dataset $D$
        \FOR{every iteration}    
            \STATE Randomly sample a batch of videos $\{o_{1:T}\}^b\sim \mathcal{D}$
            \STATE Sample a temporal correlation scale $h  \in [1, T-2]$
            \STATE Compute $\mathcal{L}_{\text{MML}}$ in Eq~\eqref{eq.6}
            \STATE Augment observations $o_{i}^{\prime } = \text{aug}\left ( o_{i} \right ) $
            \STATE Compute $\mathcal{L_{\text{SAL}}}$ in Eq~\eqref{eq.7}
            \STATE Obtain the overall context $\bar{c} = \text{mean} \left ( c_{1:T}  \right ) $
            \STATE Minimizing optimization objectives: $\mathcal{L}_{\text{MML}},\mathcal{L_{\text{SAL}}}$
            \STATE Update action-free world model
        \ENDFOR
	\end{algorithmic} 
\end{algorithm}

The pre-training process is shown in Algorithm~\ref{alg1}, we jointly optimize the MML and SAL objective functions \(\mathcal{L}_{\text{MML}}\) and \(\mathcal{L}_{\text{SAL}}\), and update the action-free world model.
During the fine-tuning phase, we also optimize the two objective functions as well as the action-conditional world model.
Please refer to Appendix 1 for details on the architecture and algorithm of the fine-tuning stage.

\begin{figure*}
  \centering
  \includegraphics[width=0.98\linewidth]{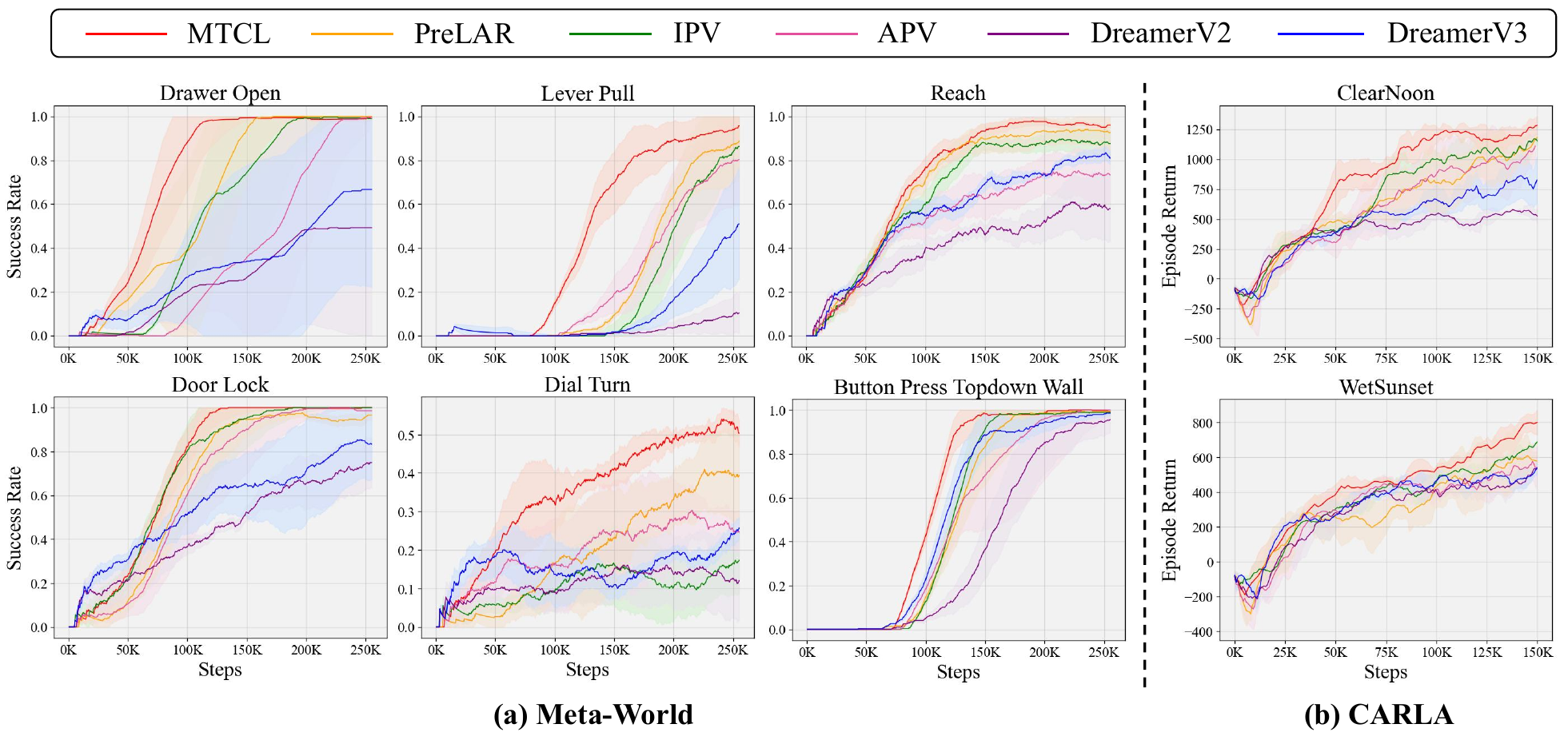}
  \caption{\textbf{Meta-World (\textit{left}) and CARLA (\textit{right}) results.}
  (a) Learning curves of MTCL (ours) compared to baselines across six tasks in Meta-World, based on the average success rate over five runs.
  (b) Learning curves of MTCL (ours) compared to baselines in the ``ClearNoon'' and ``WetSunset'' weather conditions on CARLA, measured by average episode return over five runs.}
  \label{result of Meta-World and CARLA}
\end{figure*}

\section{Experiments}

We conduct extensive experiments on three different downstream benchmarks: DMControl Remastered (DMCR)~\cite{DBLP:journals/corr/abs-2010-06740}, Meta-World~\cite{DBLP:conf/corl/YuQHJHFL19}, and CARLA~\cite{DBLP:conf/corl/DosovitskiyRCLK17}.
Our experiments investigate the following four questions:

\begin{itemize}
    \item Is the pre-training process broadly applicable to various downstream tasks?

    \item Are the proposed MML and SAL modules necessary?

    \item Does our method truly affect the pre-training phase?

    \item Does our method retain more informative representations during pre-training?
\end{itemize}

\textbf{Pre-training datasets.} 
Consistent with IPV, we use the Something-Something-V2 (SSV2) dataset~\cite{DBLP:conf/iccv/GoyalKMMWKHFYMH17} for pre-training.

\textbf{Baselines.}
We compare our method with two state-of-the-art Model-Based Reinforcement Learning (MBRL) algorithms and three unsupervised pre-training approaches in RL:
1) \textbf{DreamerV2}~\cite{DBLP:conf/iclr/HafnerL0B21} is the first MBRL algorithm to reach human-level performance on the Atari benchmark.
2) \textbf{DreamerV3}~\cite{DBLP:journals/corr/abs-2301-04104}, is currently a powerful MBRL algorithm, outperforming specialized approaches across more than 150 diverse tasks.
3) \textbf{APV}~\cite{DBLP:conf/icml/SeoLJA22} pre-trains an action-free world model using unlabeled video data from a specific domain: RLBench~\cite{DBLP:journals/ral/JamesMAD20}. 
4) \textbf{IPV}~\cite{DBLP:conf/nips/0001MDL23} introduces Contextualized World Models in pre-training with rich internet videos.
5) \textbf{PreLAR}~\cite{DBLP:conf/eccv/ZhangKSC24} introduces a learnable action representation to leverage action-free videos for pre-training world models.
Notably, DreamerV2 and DreamerV3 adopt the model-based RL paradigm, which only supports online training without pre-training.

\subsection{Evaluation on DMControl Remastered}
DMC Remastered (DMCR) is a challenging version of DeepMind Control Suite~\cite{DBLP:journals/corr/abs-1801-00690}, which is a popular simulated robotics benchmark.
DMCR features randomly generated complex graphics to measure visual generalization in continuous control. 
Following IPV, we select the same three tasks: ``Walker Run'', ``Cheetah Run'', and ``Hopper Stand''. 
As shown in Figure~\ref{result of DMControl Remastered}, we present the learning curves of our method (MTCL) compared with baselines on these tasks.
Additionally, the fourth figure in Figure~\ref{result of DMControl Remastered} shows the average performance of each algorithm across three tasks.
We run each task with five seeds. 
The solid lines represent the mean return, and the shaded areas represent the standard deviation.
Notably, our method significantly improves sample efficiency and asymptotic performance across all three tasks.

\subsection{Evaluation on Meta-World}
Meta-World is a widely used robotics benchmark with 50 different manipulation tasks.
We evaluate our algorithm on the same six tasks used for IPV and PreLAR.

Figure~\ref{result of Meta-World and CARLA} (a) shows the learning curves of our method (MTCL) compared to other baselines.
Our method achieves state-of-the-art results across all six tasks.  
Notably, while our network backbone is based on DreamerV2, it also outperforms the more powerful MBRL algorithm, DreamerV3.
It can be observed that our method improves sample efficiency across all tasks, particularly in ``Drawer Open'', ``Lever Pull'', ``Dial Turn'' and ``Button Press Topdown Wall''. 
In addition, there is a significant improvement in asymptotic performance in ``Lever Pull'' and ``Dial Turn''.
The ``Dial Turn'' is particularly challenging, but our method successfully masters it to some extent.

\subsection{Evaluation on CARLA}
CARLA is a challenging autonomous driving benchmark. 
The agent aims to maximize the distance traveled along the highway within 1000 steps while minimizing collisions.
We evaluate our method and baselines under the ``ClearNoon'' and ``WetSunset'' weather conditions, as used in IPV.
Figure~\ref{result of Meta-World and CARLA} (b) presents the learning curves of our method (MTCL) compared to other baselines.
Our method demonstrates superior sample efficiency and asymptotic performance.
Especially in the challenging ``WetSunset'' weather, our method achieves a considerable improvement in asymptotic performance.
Under the relatively simple ``ClearNoon'' weather, our method also significantly improves sample efficiency.

In summary, our method consistently improves sample efficiency and asymptotic performance across various downstream tasks. 
This demonstrates that our pre-training process, which independently models multi-scale temporal correlations rather than learning from the pixel space, helps capture more informative representations. 
As a result, it is broadly applicable to various downstream tasks.

\begin{figure*}
  \centering
  \includegraphics[width=0.795\linewidth]{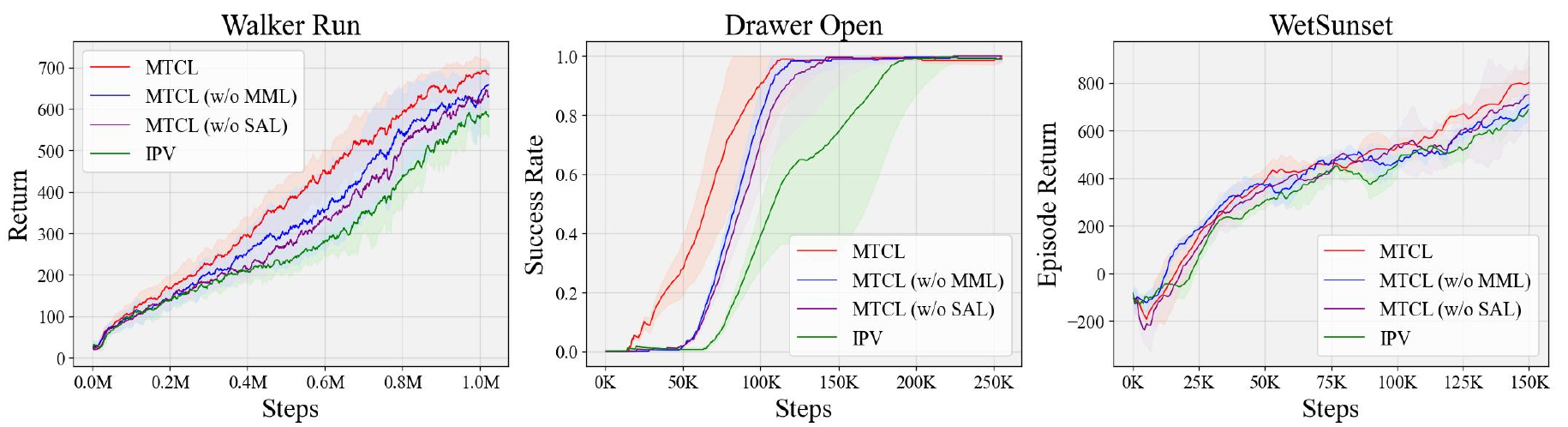}
  \caption{\textbf{Ablation Study.}  We conduct ablation studies on the MML and SAL components, selecting one task each from DMCR,
Meta-World, and CARLA.}
  \label{Ablation Study}
\end{figure*}

\begin{figure*}
  \centering
  \includegraphics[width=0.795\linewidth]{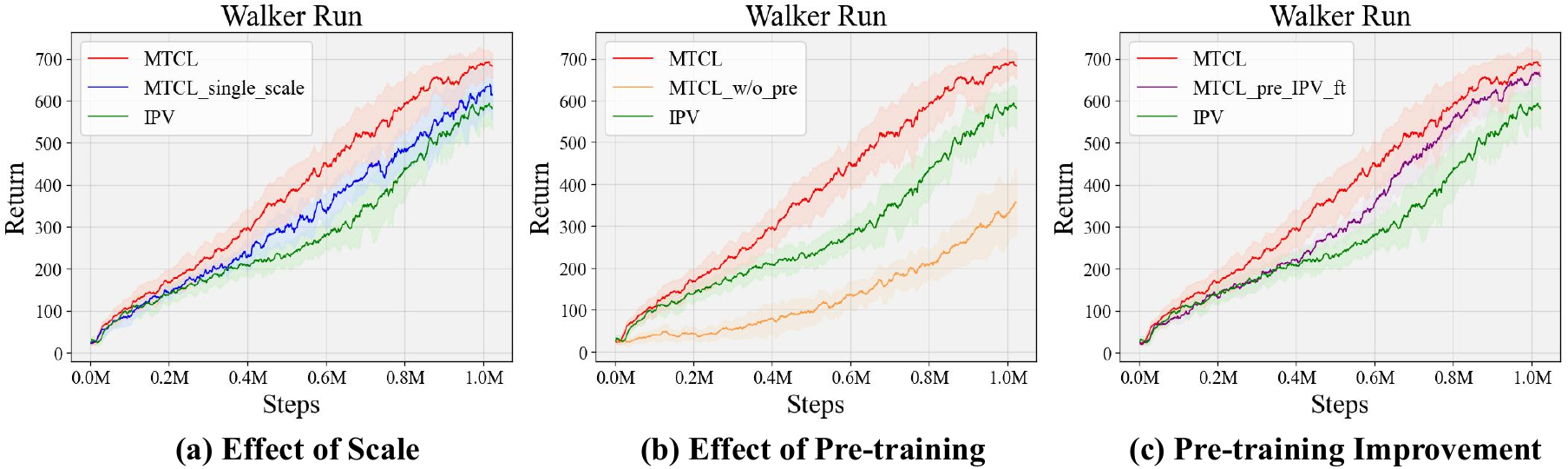}
  \caption{\textbf{Variant Experiments.}  
  (a) We compare the effectiveness of multi-scale temporal correlation modeling with single-scale modeling.
  (b) We investigate the effectiveness of the pre-training paradigm.
  (c) We compare IPV using our pre-training process with the original IPV, showing that using our pre-training representations results in better performance.
  }
  \label{Variant Experiments}
\end{figure*}

\subsection{Ablation Study}
We conduct a series of ablation studies to evaluate the effectiveness of the two components of our method (MTCL): Multi-scale Motion-aware Learning (MML) and Static Appearance-aware Learning (SAL). 
Each component is removed individually to assess its contribution to overall performance. 
We select one task from each of the three chosen downstream benchmarks—DMCR, Meta-World, and CARLA—to conduct ablation experiments.
The results are shown in Figure~\ref{Ablation Study}, where ``MTCL (w/o MML)'' represents the model without Multi-scale Motion-aware Learning, and ``MTCL (w/o SAL)'' represents the model without Static Appearance-aware Learning.
The results show that removing either component leads to a decrease in performance, demonstrating that MML and SAL are necessary and complementary. 
These two components work together to facilitate the learning of more informative representations.

\subsection{Variant Experiments}
In this section, we will explore the effects of the multi-scale modeling in our method, the impact of the pre-training paradigm, and whether our method truly improves the pre-training process.

\textbf{Single-scale Modeling.} 
To illustrate the necessity of multi-scale modeling, we perform variant experiments to explore the impact of using only a single scale.
As shown in Figure~\ref{Variant Experiments} (a), ``MTCL\_single\_scale'' refers to the variant where MML uses only the strongest temporal correlation scale.
The multi-scale approach clearly outperforms the single-scale variant, demonstrating that focusing on various temporal correlation scales helps capture more informative representations.

\textbf{Without Pre-training Paradigm.} 
To assess the effectiveness of pre-training, we define the ``MTCL\_w/o\_pre'' variant that removes the pre-training phase and learns from scratch in downstream tasks, as shown in Figure~\ref{Variant Experiments} (b).
By comparing ``MTCL'' and ``MTCL\_w/o\_pre'', we observe that the pre-training paradigm significantly improves sample efficiency and asymptotic performance in downstream tasks.

\textbf{Pretraining Improvements.}
We conduct additional experiments to show that our method truly improves the pre-training process.
In Figure~\ref{Variant Experiments} (c), ``MTCL\_pre\_IPV\_ft'' refers to IPV using our pre-training process.
Compared to the original IPV, using our pre-training process significantly improves sample efficiency and asymptotic performance. 
This indicates that our method truly affects the pre-training phase, leading to more informative representations that facilitate policy learning in downstream tasks.

\begin{figure*}
  \centering
  \includegraphics[width=0.716\linewidth]{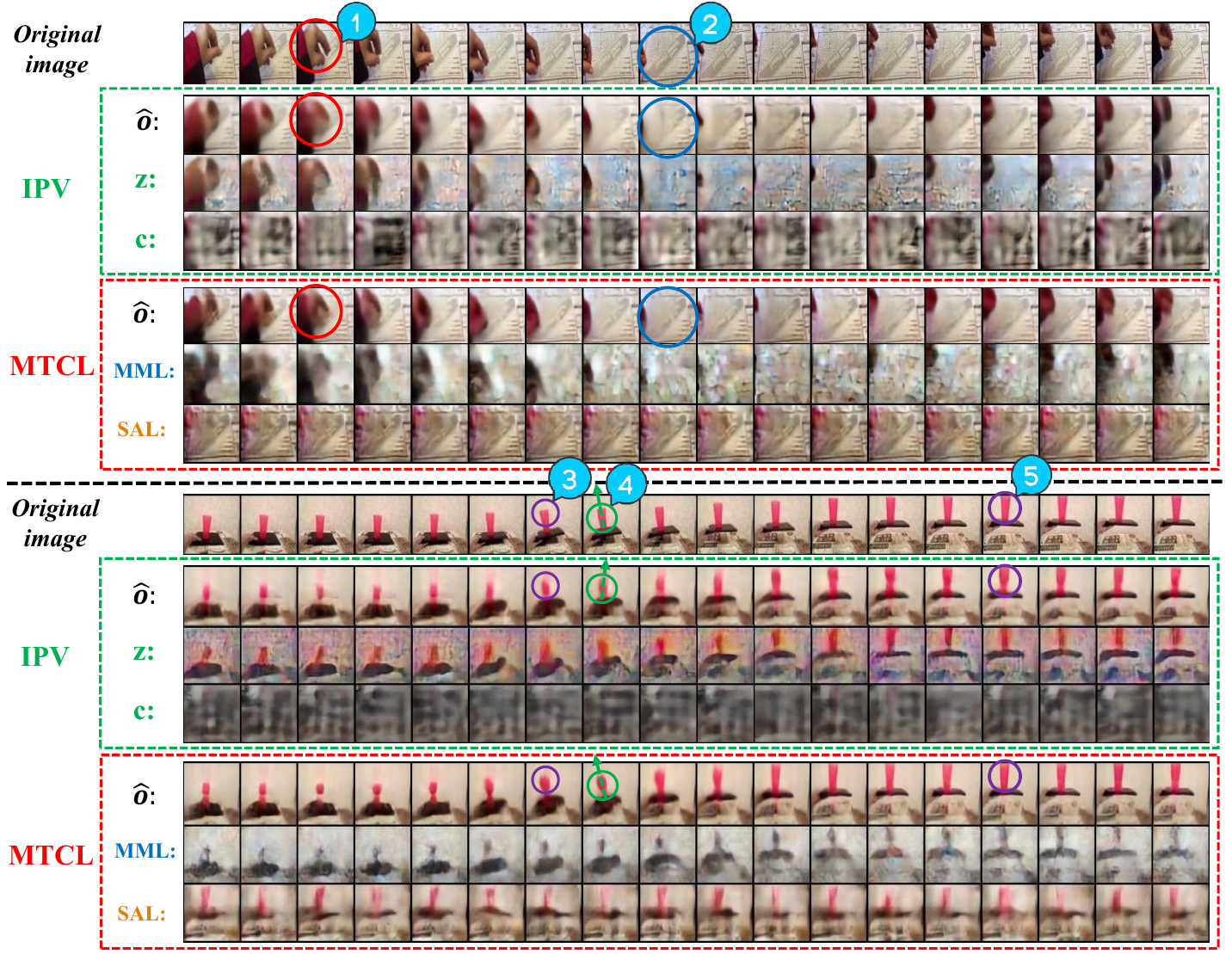}  \caption{\textbf{Visualization of Reconstruction.} We select two video examples and visualize the reconstructed images \(\hat{o}\) of our method (MTCL) and IPV, respectively. Additionally, we also visualize the features captured by the MML and SAL components of our method (MTCL), as well as the representations \(z\) and context variables \(c\) of IPV.}
  \label{Reconstruction}
\end{figure*}

\subsection{Qualitative Experiments}
To validate that our method acquires more informative representations during pre-training, we conduct feature visualizations and future frames prediction experiments.

\textbf{Visualization of Features.}
Our method effectively learns more informative representations, addressing the issue of information omission present in traditional methods. 
As shown in Figure~\ref{Reconstruction}, where \(\hat{o}\) represents the reconstructed image, our method (MTCL) successfully captures finer-grained details compared to the strongest baseline, IPV.
We illustrate the details captured by our method using five labeled examples from Figure~\ref{Reconstruction}.
Specifically, in the reconstructed image \(\hat{o}\), we observe that our method: 1) more accurately captures the shape and contours of the hand, 2) more finely captures patterns and annotations on the drawing, 3) more accurately identifies the shape of the candle's head, 4) more precisely captures the orientation of the candle, and 5) more accurately captures the shape of the candle's body, reducing deformation.

Additionally, the MML and SAL components of our method (MTCL) focus on motion and static objects, respectively. 
As shown in Figure~\ref{Reconstruction}, MML emphasizes the contours of motion objects, such as the shape of the hand or the outline of the candlestick.
SAL focuses on static background details and colors, such as the static patterns and annotations on the drawing, or the color of the candlestick and the layout of the table. 
In contrast, IPV primarily focuses on elements with large-proportion pixels, such as basic colors and textures, resulting in significant loss of detailed information.

\begin{figure}
  \centering
  \includegraphics[width=\linewidth]{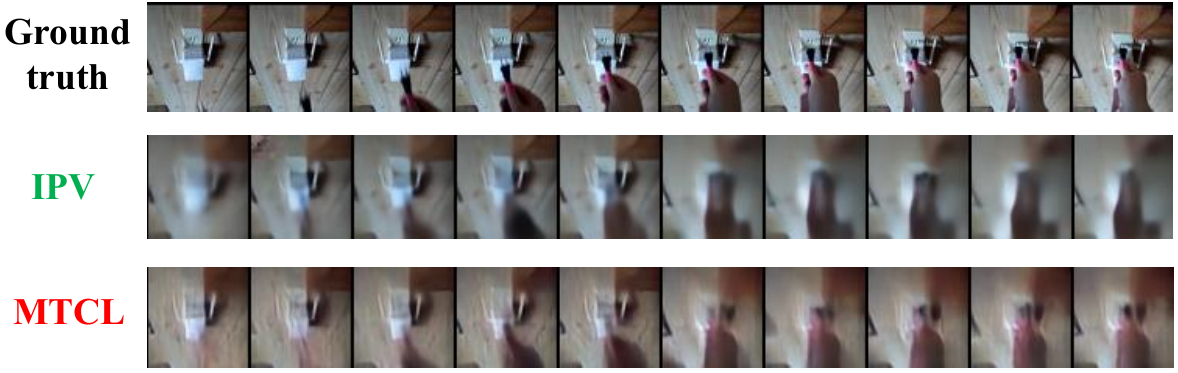}
  \caption{\textbf{Video Prediction.}  We present the predicted future frames of our method (MTCL) and IPV respectively. }
  \label{Video Prediction}
\end{figure}

\textbf{Video Prediction.}
As shown in Figure~\ref{Video Prediction}, we compare the future frames predicted by our method (MTCL) and IPV.
Unlike in reconstruction, only the first frame is real during the prediction process.
Our predictions consistently retain detailed information about the hand and plug throughout the process, while IPV's predictions eventually reduce to rough contours.
These results demonstrate that our method preserves sufficient information in the representation, enabling more accurate dynamics prediction.

\section{Conclusion}
In this paper, we address the issue with previous unsupervised pre-training methods of RL, which tend to retain large-proportion stationary information while omitting small but crucial elements.
Without additional prior knowledge about downstream tasks, paying equal attention to different elements in videos is essential.
To achieve this, we first propose a temporal correlation space where elements are inherently separable.
Then, we independently model multi-scale temporal correlations by setting a series of contrastive learning objectives.
This approach allows us to learn more informative representations.
Experimental results show that our method achieves state-of-the-art sample efficiency and asymptotic performance in various downstream tasks.

\begin{acks}
This work was supported by the Fundamental Research Funds for the Central Universities (Grant No. 2025XKBH006) and the Aeronautical Science Foundation of China (Grant No. 202300010M5001). 
\end{acks}

\bibliographystyle{ACM-Reference-Format}
\balance
\bibliography{sample-base}


\begin{thebibliography}{41}


\ifx \showCODEN    \undefined \def \showCODEN     #1{\unskip}     \fi
\ifx \showISBNx    \undefined \def \showISBNx     #1{\unskip}     \fi
\ifx \showISBNxiii \undefined \def \showISBNxiii  #1{\unskip}     \fi
\ifx \showISSN     \undefined \def \showISSN      #1{\unskip}     \fi
\ifx \showLCCN     \undefined \def \showLCCN      #1{\unskip}     \fi
\ifx \shownote     \undefined \def \shownote      #1{#1}          \fi
\ifx \showarticletitle \undefined \def \showarticletitle #1{#1}   \fi
\ifx \showURL      \undefined \def \showURL       {\relax}        \fi
\providecommand\bibfield[2]{#2}
\providecommand\bibinfo[2]{#2}
\providecommand\natexlab[1]{#1}
\providecommand\showeprint[2][]{arXiv:#2}

\bibitem[Blei et~al\mbox{.}(2016)]%
        {DBLP:journals/corr/BleiKM16}
\bibfield{author}{\bibinfo{person}{David~M. Blei}, \bibinfo{person}{Alp Kucukelbir}, {and} \bibinfo{person}{Jon~D. McAuliffe}.} \bibinfo{year}{2016}\natexlab{}.
\newblock \showarticletitle{Variational Inference: {A} Review for Statisticians}.
\newblock \bibinfo{journal}{\emph{CoRR}}  \bibinfo{volume}{abs/1601.00670} (\bibinfo{year}{2016}).
\newblock
\showeprint[arXiv]{1601.00670}
\urldef\tempurl%
\url{http://arxiv.org/abs/1601.00670}
\showURL{%
\tempurl}


\bibitem[Brown et~al\mbox{.}(2020)]%
        {DBLP:conf/nips/BrownMRSKDNSSAA20}
\bibfield{author}{\bibinfo{person}{Tom~B. Brown}, \bibinfo{person}{Benjamin Mann}, \bibinfo{person}{Nick Ryder}, \bibinfo{person}{Melanie Subbiah}, \bibinfo{person}{Jared Kaplan}, \bibinfo{person}{Prafulla Dhariwal}, \bibinfo{person}{Arvind Neelakantan}, \bibinfo{person}{Pranav Shyam}, \bibinfo{person}{Girish Sastry}, \bibinfo{person}{Amanda Askell}, \bibinfo{person}{Sandhini Agarwal}, \bibinfo{person}{Ariel Herbert{-}Voss}, \bibinfo{person}{Gretchen Krueger}, \bibinfo{person}{Tom Henighan}, \bibinfo{person}{Rewon Child}, \bibinfo{person}{Aditya Ramesh}, \bibinfo{person}{Daniel~M. Ziegler}, \bibinfo{person}{Jeffrey Wu}, \bibinfo{person}{Clemens Winter}, \bibinfo{person}{Christopher Hesse}, \bibinfo{person}{Mark Chen}, \bibinfo{person}{Eric Sigler}, \bibinfo{person}{Mateusz Litwin}, \bibinfo{person}{Scott Gray}, \bibinfo{person}{Benjamin Chess}, \bibinfo{person}{Jack Clark}, \bibinfo{person}{Christopher Berner}, \bibinfo{person}{Sam McCandlish}, \bibinfo{person}{Alec Radford}, \bibinfo{person}{Ilya Sutskever}, {and} \bibinfo{person}{Dario Amodei}.} \bibinfo{year}{2020}\natexlab{}.
\newblock \showarticletitle{Language Models are Few-Shot Learners}. In \bibinfo{booktitle}{\emph{Advances in Neural Information Processing Systems 33: Annual Conference on Neural Information Processing Systems 2020, NeurIPS 2020, December 6-12, 2020, virtual}}, \bibfield{editor}{\bibinfo{person}{Hugo Larochelle}, \bibinfo{person}{Marc'Aurelio Ranzato}, \bibinfo{person}{Raia Hadsell}, \bibinfo{person}{Maria{-}Florina Balcan}, {and} \bibinfo{person}{Hsuan{-}Tien Lin}} (Eds.).
\newblock
\urldef\tempurl%
\url{https://proceedings.neurips.cc/paper/2020/hash/1457c0d6bfcb4967418bfb8ac142f64a-Abstract.html}
\showURL{%
\tempurl}


\bibitem[Chen et~al\mbox{.}(2022)]%
        {DBLP:journals/corr/abs-2202-09481}
\bibfield{author}{\bibinfo{person}{Chang Chen}, \bibinfo{person}{Yi{-}Fu Wu}, \bibinfo{person}{Jaesik Yoon}, {and} \bibinfo{person}{Sungjin Ahn}.} \bibinfo{year}{2022}\natexlab{}.
\newblock \showarticletitle{TransDreamer: Reinforcement Learning with Transformer World Models}.
\newblock \bibinfo{journal}{\emph{CoRR}}  \bibinfo{volume}{abs/2202.09481} (\bibinfo{year}{2022}).
\newblock
\showeprint[arXiv]{2202.09481}
\urldef\tempurl%
\url{https://arxiv.org/abs/2202.09481}
\showURL{%
\tempurl}


\bibitem[Dedieu et~al\mbox{.}(2025)]%
        {DBLP:journals/corr/abs-2502-01591}
\bibfield{author}{\bibinfo{person}{Antoine Dedieu}, \bibinfo{person}{Joseph Ortiz}, \bibinfo{person}{Xinghua Lou}, \bibinfo{person}{Carter Wendelken}, \bibinfo{person}{Wolfgang Lehrach}, \bibinfo{person}{J.~Swaroop Guntupalli}, \bibinfo{person}{Miguel L{\'{a}}zaro{-}Gredilla}, {and} \bibinfo{person}{Kevin~Patrick Murphy}.} \bibinfo{year}{2025}\natexlab{}.
\newblock \showarticletitle{Improving Transformer World Models for Data-Efficient {RL}}.
\newblock \bibinfo{journal}{\emph{CoRR}}  \bibinfo{volume}{abs/2502.01591} (\bibinfo{year}{2025}).
\newblock
\showeprint[arXiv]{2502.01591}
\href{https://doi.org/10.48550/ARXIV.2502.01591}{doi:\nolinkurl{10.48550/ARXIV.2502.01591}}


\bibitem[Devlin et~al\mbox{.}(2019)]%
        {DBLP:conf/naacl/DevlinCLT19}
\bibfield{author}{\bibinfo{person}{Jacob Devlin}, \bibinfo{person}{Ming{-}Wei Chang}, \bibinfo{person}{Kenton Lee}, {and} \bibinfo{person}{Kristina Toutanova}.} \bibinfo{year}{2019}\natexlab{}.
\newblock \showarticletitle{{BERT:} Pre-training of Deep Bidirectional Transformers for Language Understanding}. In \bibinfo{booktitle}{\emph{Proceedings of the 2019 Conference of the North American Chapter of the Association for Computational Linguistics: Human Language Technologies, {NAACL-HLT} 2019, Minneapolis, MN, USA, June 2-7, 2019, Volume 1 (Long and Short Papers)}}, \bibfield{editor}{\bibinfo{person}{Jill Burstein}, \bibinfo{person}{Christy Doran}, {and} \bibinfo{person}{Thamar Solorio}} (Eds.). \bibinfo{publisher}{Association for Computational Linguistics}, \bibinfo{pages}{4171--4186}.
\newblock
\href{https://doi.org/10.18653/V1/N19-1423}{doi:\nolinkurl{10.18653/V1/N19-1423}}


\bibitem[Dosovitskiy et~al\mbox{.}(2021)]%
        {DBLP:conf/iclr/DosovitskiyB0WZ21}
\bibfield{author}{\bibinfo{person}{Alexey Dosovitskiy}, \bibinfo{person}{Lucas Beyer}, \bibinfo{person}{Alexander Kolesnikov}, \bibinfo{person}{Dirk Weissenborn}, \bibinfo{person}{Xiaohua Zhai}, \bibinfo{person}{Thomas Unterthiner}, \bibinfo{person}{Mostafa Dehghani}, \bibinfo{person}{Matthias Minderer}, \bibinfo{person}{Georg Heigold}, \bibinfo{person}{Sylvain Gelly}, \bibinfo{person}{Jakob Uszkoreit}, {and} \bibinfo{person}{Neil Houlsby}.} \bibinfo{year}{2021}\natexlab{}.
\newblock \showarticletitle{An Image is Worth 16x16 Words: Transformers for Image Recognition at Scale}. In \bibinfo{booktitle}{\emph{9th International Conference on Learning Representations, {ICLR} 2021, Virtual Event, Austria, May 3-7, 2021}}. \bibinfo{publisher}{OpenReview.net}.
\newblock
\urldef\tempurl%
\url{https://openreview.net/forum?id=YicbFdNTTy}
\showURL{%
\tempurl}


\bibitem[Dosovitskiy et~al\mbox{.}(2017)]%
        {DBLP:conf/corl/DosovitskiyRCLK17}
\bibfield{author}{\bibinfo{person}{Alexey Dosovitskiy}, \bibinfo{person}{Germ{\'{a}}n Ros}, \bibinfo{person}{Felipe Codevilla}, \bibinfo{person}{Antonio~M. L{\'{o}}pez}, {and} \bibinfo{person}{Vladlen Koltun}.} \bibinfo{year}{2017}\natexlab{}.
\newblock \showarticletitle{{CARLA:} An Open Urban Driving Simulator}. In \bibinfo{booktitle}{\emph{1st Annual Conference on Robot Learning, CoRL 2017, Mountain View, California, USA, November 13-15, 2017, Proceedings}} \emph{(\bibinfo{series}{Proceedings of Machine Learning Research}, Vol.~\bibinfo{volume}{78})}. \bibinfo{publisher}{{PMLR}}, \bibinfo{pages}{1--16}.
\newblock
\urldef\tempurl%
\url{http://proceedings.mlr.press/v78/dosovitskiy17a.html}
\showURL{%
\tempurl}


\bibitem[Ghosh et~al\mbox{.}(2023)]%
        {DBLP:conf/icml/GhoshBL23}
\bibfield{author}{\bibinfo{person}{Dibya Ghosh}, \bibinfo{person}{Chethan~Anand Bhateja}, {and} \bibinfo{person}{Sergey Levine}.} \bibinfo{year}{2023}\natexlab{}.
\newblock \showarticletitle{Reinforcement Learning from Passive Data via Latent Intentions}. In \bibinfo{booktitle}{\emph{International Conference on Machine Learning, {ICML} 2023, 23-29 July 2023, Honolulu, Hawaii, {USA}}} \emph{(\bibinfo{series}{Proceedings of Machine Learning Research}, Vol.~\bibinfo{volume}{202})}, \bibfield{editor}{\bibinfo{person}{Andreas Krause}, \bibinfo{person}{Emma Brunskill}, \bibinfo{person}{Kyunghyun Cho}, \bibinfo{person}{Barbara Engelhardt}, \bibinfo{person}{Sivan Sabato}, {and} \bibinfo{person}{Jonathan Scarlett}} (Eds.). \bibinfo{publisher}{{PMLR}}, \bibinfo{pages}{11321--11339}.
\newblock
\urldef\tempurl%
\url{https://proceedings.mlr.press/v202/ghosh23a.html}
\showURL{%
\tempurl}


\bibitem[Goyal et~al\mbox{.}(2017)]%
        {DBLP:conf/iccv/GoyalKMMWKHFYMH17}
\bibfield{author}{\bibinfo{person}{Raghav Goyal}, \bibinfo{person}{Samira~Ebrahimi Kahou}, \bibinfo{person}{Vincent Michalski}, \bibinfo{person}{Joanna Materzynska}, \bibinfo{person}{Susanne Westphal}, \bibinfo{person}{Heuna Kim}, \bibinfo{person}{Valentin Haenel}, \bibinfo{person}{Ingo Fr{\"{u}}nd}, \bibinfo{person}{Peter Yianilos}, \bibinfo{person}{Moritz Mueller{-}Freitag}, \bibinfo{person}{Florian Hoppe}, \bibinfo{person}{Christian Thurau}, \bibinfo{person}{Ingo Bax}, {and} \bibinfo{person}{Roland Memisevic}.} \bibinfo{year}{2017}\natexlab{}.
\newblock \showarticletitle{The "Something Something" Video Database for Learning and Evaluating Visual Common Sense}. In \bibinfo{booktitle}{\emph{{IEEE} International Conference on Computer Vision, {ICCV} 2017, Venice, Italy, October 22-29, 2017}}. \bibinfo{publisher}{{IEEE} Computer Society}, \bibinfo{pages}{5843--5851}.
\newblock
\href{https://doi.org/10.1109/ICCV.2017.622}{doi:\nolinkurl{10.1109/ICCV.2017.622}}


\bibitem[Grigsby and Qi(2020)]%
        {DBLP:journals/corr/abs-2010-06740}
\bibfield{author}{\bibinfo{person}{Jake Grigsby} {and} \bibinfo{person}{Yanjun Qi}.} \bibinfo{year}{2020}\natexlab{}.
\newblock \showarticletitle{Measuring Visual Generalization in Continuous Control from Pixels}.
\newblock \bibinfo{journal}{\emph{CoRR}}  \bibinfo{volume}{abs/2010.06740} (\bibinfo{year}{2020}).
\newblock
\showeprint[arXiv]{2010.06740}
\urldef\tempurl%
\url{https://arxiv.org/abs/2010.06740}
\showURL{%
\tempurl}


\bibitem[Hafner et~al\mbox{.}(2020)]%
        {DBLP:conf/iclr/HafnerLB020}
\bibfield{author}{\bibinfo{person}{Danijar Hafner}, \bibinfo{person}{Timothy~P. Lillicrap}, \bibinfo{person}{Jimmy Ba}, {and} \bibinfo{person}{Mohammad Norouzi}.} \bibinfo{year}{2020}\natexlab{}.
\newblock \showarticletitle{Dream to Control: Learning Behaviors by Latent Imagination}. In \bibinfo{booktitle}{\emph{8th International Conference on Learning Representations, {ICLR} 2020, Addis Ababa, Ethiopia, April 26-30, 2020}}. \bibinfo{publisher}{OpenReview.net}.
\newblock
\urldef\tempurl%
\url{https://openreview.net/forum?id=S1lOTC4tDS}
\showURL{%
\tempurl}


\bibitem[Hafner et~al\mbox{.}(2019)]%
        {DBLP:conf/icml/HafnerLFVHLD19}
\bibfield{author}{\bibinfo{person}{Danijar Hafner}, \bibinfo{person}{Timothy~P. Lillicrap}, \bibinfo{person}{Ian Fischer}, \bibinfo{person}{Ruben Villegas}, \bibinfo{person}{David Ha}, \bibinfo{person}{Honglak Lee}, {and} \bibinfo{person}{James Davidson}.} \bibinfo{year}{2019}\natexlab{}.
\newblock \showarticletitle{Learning Latent Dynamics for Planning from Pixels}. In \bibinfo{booktitle}{\emph{Proceedings of the 36th International Conference on Machine Learning, {ICML} 2019, 9-15 June 2019, Long Beach, California, {USA}}} \emph{(\bibinfo{series}{Proceedings of Machine Learning Research}, Vol.~\bibinfo{volume}{97})}, \bibfield{editor}{\bibinfo{person}{Kamalika Chaudhuri} {and} \bibinfo{person}{Ruslan Salakhutdinov}} (Eds.). \bibinfo{publisher}{{PMLR}}, \bibinfo{pages}{2555--2565}.
\newblock
\urldef\tempurl%
\url{http://proceedings.mlr.press/v97/hafner19a.html}
\showURL{%
\tempurl}


\bibitem[Hafner et~al\mbox{.}(2021)]%
        {DBLP:conf/iclr/HafnerL0B21}
\bibfield{author}{\bibinfo{person}{Danijar Hafner}, \bibinfo{person}{Timothy~P. Lillicrap}, \bibinfo{person}{Mohammad Norouzi}, {and} \bibinfo{person}{Jimmy Ba}.} \bibinfo{year}{2021}\natexlab{}.
\newblock \showarticletitle{Mastering Atari with Discrete World Models}. In \bibinfo{booktitle}{\emph{9th International Conference on Learning Representations, {ICLR} 2021, Virtual Event, Austria, May 3-7, 2021}}. \bibinfo{publisher}{OpenReview.net}.
\newblock
\urldef\tempurl%
\url{https://openreview.net/forum?id=0oabwyZbOu}
\showURL{%
\tempurl}


\bibitem[Hafner et~al\mbox{.}(2023)]%
        {DBLP:journals/corr/abs-2301-04104}
\bibfield{author}{\bibinfo{person}{Danijar Hafner}, \bibinfo{person}{Jurgis Pasukonis}, \bibinfo{person}{Jimmy Ba}, {and} \bibinfo{person}{Timothy~P. Lillicrap}.} \bibinfo{year}{2023}\natexlab{}.
\newblock \showarticletitle{Mastering Diverse Domains through World Models}.
\newblock \bibinfo{journal}{\emph{CoRR}}  \bibinfo{volume}{abs/2301.04104} (\bibinfo{year}{2023}).
\newblock


\bibitem[Hansen et~al\mbox{.}(2021a)]%
        {DBLP:conf/iclr/HansenJSAAEPW21}
\bibfield{author}{\bibinfo{person}{Nicklas Hansen}, \bibinfo{person}{Rishabh Jangir}, \bibinfo{person}{Yu Sun}, \bibinfo{person}{Guillem Aleny{\`{a}}}, \bibinfo{person}{Pieter Abbeel}, \bibinfo{person}{Alexei~A. Efros}, \bibinfo{person}{Lerrel Pinto}, {and} \bibinfo{person}{Xiaolong Wang}.} \bibinfo{year}{2021}\natexlab{a}.
\newblock \showarticletitle{Self-Supervised Policy Adaptation during Deployment}. In \bibinfo{booktitle}{\emph{9th International Conference on Learning Representations, {ICLR} 2021, Virtual Event, Austria, May 3-7, 2021}}. \bibinfo{publisher}{OpenReview.net}.
\newblock
\urldef\tempurl%
\url{https://openreview.net/forum?id=o\_V-MjyyGV\_}
\showURL{%
\tempurl}


\bibitem[Hansen et~al\mbox{.}(2021b)]%
        {DBLP:conf/nips/HansenSW21}
\bibfield{author}{\bibinfo{person}{Nicklas Hansen}, \bibinfo{person}{Hao Su}, {and} \bibinfo{person}{Xiaolong Wang}.} \bibinfo{year}{2021}\natexlab{b}.
\newblock \showarticletitle{Stabilizing Deep Q-Learning with ConvNets and Vision Transformers under Data Augmentation}. In \bibinfo{booktitle}{\emph{Advances in Neural Information Processing Systems 34: Annual Conference on Neural Information Processing Systems 2021, NeurIPS 2021, December 6-14, 2021, virtual}}, \bibfield{editor}{\bibinfo{person}{Marc'Aurelio Ranzato}, \bibinfo{person}{Alina Beygelzimer}, \bibinfo{person}{Yann~N. Dauphin}, \bibinfo{person}{Percy Liang}, {and} \bibinfo{person}{Jennifer~Wortman Vaughan}} (Eds.). \bibinfo{pages}{3680--3693}.
\newblock
\urldef\tempurl%
\url{https://proceedings.neurips.cc/paper/2021/hash/1e0f65eb20acbfb27ee05ddc000b50ec-Abstract.html}
\showURL{%
\tempurl}


\bibitem[Hansen et~al\mbox{.}(2022)]%
        {DBLP:conf/icml/HansenSW22}
\bibfield{author}{\bibinfo{person}{Nicklas Hansen}, \bibinfo{person}{Hao Su}, {and} \bibinfo{person}{Xiaolong Wang}.} \bibinfo{year}{2022}\natexlab{}.
\newblock \showarticletitle{Temporal Difference Learning for Model Predictive Control}. In \bibinfo{booktitle}{\emph{International Conference on Machine Learning, {ICML} 2022, 17-23 July 2022, Baltimore, Maryland, {USA}}} \emph{(\bibinfo{series}{Proceedings of Machine Learning Research}, Vol.~\bibinfo{volume}{162})}, \bibfield{editor}{\bibinfo{person}{Kamalika Chaudhuri}, \bibinfo{person}{Stefanie Jegelka}, \bibinfo{person}{Le~Song}, \bibinfo{person}{Csaba Szepesv{\'{a}}ri}, \bibinfo{person}{Gang Niu}, {and} \bibinfo{person}{Sivan Sabato}} (Eds.). \bibinfo{publisher}{{PMLR}}, \bibinfo{pages}{8387--8406}.
\newblock


\bibitem[He et~al\mbox{.}(2022)]%
        {DBLP:conf/cvpr/HeCXLDG22}
\bibfield{author}{\bibinfo{person}{Kaiming He}, \bibinfo{person}{Xinlei Chen}, \bibinfo{person}{Saining Xie}, \bibinfo{person}{Yanghao Li}, \bibinfo{person}{Piotr Doll{\'{a}}r}, {and} \bibinfo{person}{Ross~B. Girshick}.} \bibinfo{year}{2022}\natexlab{}.
\newblock \showarticletitle{Masked Autoencoders Are Scalable Vision Learners}. In \bibinfo{booktitle}{\emph{{IEEE/CVF} Conference on Computer Vision and Pattern Recognition, {CVPR} 2022, New Orleans, LA, USA, June 18-24, 2022}}. \bibinfo{publisher}{{IEEE}}, \bibinfo{pages}{15979--15988}.
\newblock
\href{https://doi.org/10.1109/CVPR52688.2022.01553}{doi:\nolinkurl{10.1109/CVPR52688.2022.01553}}


\bibitem[He et~al\mbox{.}(2020)]%
        {DBLP:conf/cvpr/He0WXG20}
\bibfield{author}{\bibinfo{person}{Kaiming He}, \bibinfo{person}{Haoqi Fan}, \bibinfo{person}{Yuxin Wu}, \bibinfo{person}{Saining Xie}, {and} \bibinfo{person}{Ross~B. Girshick}.} \bibinfo{year}{2020}\natexlab{}.
\newblock \showarticletitle{Momentum Contrast for Unsupervised Visual Representation Learning}. In \bibinfo{booktitle}{\emph{2020 {IEEE/CVF} Conference on Computer Vision and Pattern Recognition, {CVPR} 2020, Seattle, WA, USA, June 13-19, 2020}}. \bibinfo{publisher}{Computer Vision Foundation / {IEEE}}, \bibinfo{pages}{9726--9735}.
\newblock
\href{https://doi.org/10.1109/CVPR42600.2020.00975}{doi:\nolinkurl{10.1109/CVPR42600.2020.00975}}


\bibitem[He et~al\mbox{.}(2016)]%
        {DBLP:conf/cvpr/HeZRS16}
\bibfield{author}{\bibinfo{person}{Kaiming He}, \bibinfo{person}{Xiangyu Zhang}, \bibinfo{person}{Shaoqing Ren}, {and} \bibinfo{person}{Jian Sun}.} \bibinfo{year}{2016}\natexlab{}.
\newblock \showarticletitle{Deep Residual Learning for Image Recognition}. In \bibinfo{booktitle}{\emph{2016 {IEEE} Conference on Computer Vision and Pattern Recognition, {CVPR} 2016, Las Vegas, NV, USA, June 27-30, 2016}}. \bibinfo{publisher}{{IEEE} Computer Society}, \bibinfo{pages}{770--778}.
\newblock
\href{https://doi.org/10.1109/CVPR.2016.90}{doi:\nolinkurl{10.1109/CVPR.2016.90}}


\bibitem[Huang et~al\mbox{.}(2023)]%
        {DBLP:conf/corl/HuangWZL0023}
\bibfield{author}{\bibinfo{person}{Wenlong Huang}, \bibinfo{person}{Chen Wang}, \bibinfo{person}{Ruohan Zhang}, \bibinfo{person}{Yunzhu Li}, \bibinfo{person}{Jiajun Wu}, {and} \bibinfo{person}{Li Fei{-}Fei}.} \bibinfo{year}{2023}\natexlab{}.
\newblock \showarticletitle{VoxPoser: Composable 3D Value Maps for Robotic Manipulation with Language Models}. In \bibinfo{booktitle}{\emph{Conference on Robot Learning, CoRL 2023, 6-9 November 2023, Atlanta, GA, {USA}}} \emph{(\bibinfo{series}{Proceedings of Machine Learning Research}, Vol.~\bibinfo{volume}{229})}, \bibfield{editor}{\bibinfo{person}{Jie Tan}, \bibinfo{person}{Marc Toussaint}, {and} \bibinfo{person}{Kourosh Darvish}} (Eds.). \bibinfo{publisher}{{PMLR}}, \bibinfo{pages}{540--562}.
\newblock
\urldef\tempurl%
\url{https://proceedings.mlr.press/v229/huang23b.html}
\showURL{%
\tempurl}


\bibitem[James et~al\mbox{.}(2020)]%
        {DBLP:journals/ral/JamesMAD20}
\bibfield{author}{\bibinfo{person}{Stephen James}, \bibinfo{person}{Zicong Ma}, \bibinfo{person}{David~Rovick Arrojo}, {and} \bibinfo{person}{Andrew~J. Davison}.} \bibinfo{year}{2020}\natexlab{}.
\newblock \showarticletitle{RLBench: The Robot Learning Benchmark {\&} Learning Environment}.
\newblock \bibinfo{journal}{\emph{{IEEE} Robotics Autom. Lett.}} \bibinfo{volume}{5}, \bibinfo{number}{2} (\bibinfo{year}{2020}), \bibinfo{pages}{3019--3026}.
\newblock
\href{https://doi.org/10.1109/LRA.2020.2974707}{doi:\nolinkurl{10.1109/LRA.2020.2974707}}


\bibitem[Levine et~al\mbox{.}(2016)]%
        {DBLP:journals/jmlr/LevineFDA16}
\bibfield{author}{\bibinfo{person}{Sergey Levine}, \bibinfo{person}{Chelsea Finn}, \bibinfo{person}{Trevor Darrell}, {and} \bibinfo{person}{Pieter Abbeel}.} \bibinfo{year}{2016}\natexlab{}.
\newblock \showarticletitle{End-to-End Training of Deep Visuomotor Policies}.
\newblock \bibinfo{journal}{\emph{J. Mach. Learn. Res.}}  \bibinfo{volume}{17} (\bibinfo{year}{2016}), \bibinfo{pages}{39:1--39:40}.
\newblock
\urldef\tempurl%
\url{https://jmlr.org/papers/v17/15-522.html}
\showURL{%
\tempurl}


\bibitem[Misra et~al\mbox{.}(2024)]%
        {DBLP:journals/corr/abs-2403-13765}
\bibfield{author}{\bibinfo{person}{Dipendra Misra}, \bibinfo{person}{Akanksha Saran}, \bibinfo{person}{Tengyang Xie}, \bibinfo{person}{Alex Lamb}, {and} \bibinfo{person}{John Langford}.} \bibinfo{year}{2024}\natexlab{}.
\newblock \showarticletitle{Towards Principled Representation Learning from Videos for Reinforcement Learning}.
\newblock \bibinfo{journal}{\emph{CoRR}}  \bibinfo{volume}{abs/2403.13765} (\bibinfo{year}{2024}).
\newblock
\showeprint[arXiv]{2403.13765}
\href{https://doi.org/10.48550/ARXIV.2403.13765}{doi:\nolinkurl{10.48550/ARXIV.2403.13765}}


\bibitem[Nair et~al\mbox{.}(2022)]%
        {DBLP:conf/corl/NairRKF022}
\bibfield{author}{\bibinfo{person}{Suraj Nair}, \bibinfo{person}{Aravind Rajeswaran}, \bibinfo{person}{Vikash Kumar}, \bibinfo{person}{Chelsea Finn}, {and} \bibinfo{person}{Abhinav Gupta}.} \bibinfo{year}{2022}\natexlab{}.
\newblock \showarticletitle{{R3M:} {A} Universal Visual Representation for Robot Manipulation}. In \bibinfo{booktitle}{\emph{Conference on Robot Learning, CoRL 2022, 14-18 December 2022, Auckland, New Zealand}} \emph{(\bibinfo{series}{Proceedings of Machine Learning Research}, Vol.~\bibinfo{volume}{205})}, \bibfield{editor}{\bibinfo{person}{Karen Liu}, \bibinfo{person}{Dana Kulic}, {and} \bibinfo{person}{Jeffrey Ichnowski}} (Eds.). \bibinfo{publisher}{{PMLR}}, \bibinfo{pages}{892--909}.
\newblock
\urldef\tempurl%
\url{https://proceedings.mlr.press/v205/nair23a.html}
\showURL{%
\tempurl}


\bibitem[Okada and Taniguchi(2021)]%
        {DBLP:conf/icra/OkadaT21}
\bibfield{author}{\bibinfo{person}{Masashi Okada} {and} \bibinfo{person}{Tadahiro Taniguchi}.} \bibinfo{year}{2021}\natexlab{}.
\newblock \showarticletitle{Dreaming: Model-based Reinforcement Learning by Latent Imagination without Reconstruction}. In \bibinfo{booktitle}{\emph{{IEEE} International Conference on Robotics and Automation, {ICRA} 2021, Xi'an, China, May 30 - June 5, 2021}}. \bibinfo{publisher}{{IEEE}}, \bibinfo{pages}{4209--4215}.
\newblock
\href{https://doi.org/10.1109/ICRA48506.2021.9560734}{doi:\nolinkurl{10.1109/ICRA48506.2021.9560734}}


\bibitem[Radosavovic et~al\mbox{.}(2022)]%
        {DBLP:conf/corl/RadosavovicXJAM22}
\bibfield{author}{\bibinfo{person}{Ilija Radosavovic}, \bibinfo{person}{Tete Xiao}, \bibinfo{person}{Stephen James}, \bibinfo{person}{Pieter Abbeel}, \bibinfo{person}{Jitendra Malik}, {and} \bibinfo{person}{Trevor Darrell}.} \bibinfo{year}{2022}\natexlab{}.
\newblock \showarticletitle{Real-World Robot Learning with Masked Visual Pre-training}. In \bibinfo{booktitle}{\emph{Conference on Robot Learning, CoRL 2022, 14-18 December 2022, Auckland, New Zealand}} \emph{(\bibinfo{series}{Proceedings of Machine Learning Research}, Vol.~\bibinfo{volume}{205})}, \bibfield{editor}{\bibinfo{person}{Karen Liu}, \bibinfo{person}{Dana Kulic}, {and} \bibinfo{person}{Jeffrey Ichnowski}} (Eds.). \bibinfo{publisher}{{PMLR}}, \bibinfo{pages}{416--426}.
\newblock
\urldef\tempurl%
\url{https://proceedings.mlr.press/v205/radosavovic23a.html}
\showURL{%
\tempurl}


\bibitem[Seo et~al\mbox{.}(2022)]%
        {DBLP:conf/icml/SeoLJA22}
\bibfield{author}{\bibinfo{person}{Younggyo Seo}, \bibinfo{person}{Kimin Lee}, \bibinfo{person}{Stephen~L. James}, {and} \bibinfo{person}{Pieter Abbeel}.} \bibinfo{year}{2022}\natexlab{}.
\newblock \showarticletitle{Reinforcement Learning with Action-Free Pre-Training from Videos}. In \bibinfo{booktitle}{\emph{International Conference on Machine Learning, {ICML} 2022, 17-23 July 2022, Baltimore, Maryland, {USA}}} \emph{(\bibinfo{series}{Proceedings of Machine Learning Research}, Vol.~\bibinfo{volume}{162})}, \bibfield{editor}{\bibinfo{person}{Kamalika Chaudhuri}, \bibinfo{person}{Stefanie Jegelka}, \bibinfo{person}{Le~Song}, \bibinfo{person}{Csaba Szepesv{\'{a}}ri}, \bibinfo{person}{Gang Niu}, {and} \bibinfo{person}{Sivan Sabato}} (Eds.). \bibinfo{publisher}{{PMLR}}, \bibinfo{pages}{19561--19579}.
\newblock
\urldef\tempurl%
\url{https://proceedings.mlr.press/v162/seo22a.html}
\showURL{%
\tempurl}


\bibitem[Tassa et~al\mbox{.}(2018)]%
        {DBLP:journals/corr/abs-1801-00690}
\bibfield{author}{\bibinfo{person}{Yuval Tassa}, \bibinfo{person}{Yotam Doron}, \bibinfo{person}{Alistair Muldal}, \bibinfo{person}{Tom Erez}, \bibinfo{person}{Yazhe Li}, \bibinfo{person}{Diego de Las~Casas}, \bibinfo{person}{David Budden}, \bibinfo{person}{Abbas Abdolmaleki}, \bibinfo{person}{Josh Merel}, \bibinfo{person}{Andrew Lefrancq}, \bibinfo{person}{Timothy~P. Lillicrap}, {and} \bibinfo{person}{Martin~A. Riedmiller}.} \bibinfo{year}{2018}\natexlab{}.
\newblock \showarticletitle{DeepMind Control Suite}.
\newblock \bibinfo{journal}{\emph{CoRR}}  \bibinfo{volume}{abs/1801.00690} (\bibinfo{year}{2018}).
\newblock
\showeprint[arXiv]{1801.00690}
\urldef\tempurl%
\url{http://arxiv.org/abs/1801.00690}
\showURL{%
\tempurl}


\bibitem[Vaswani et~al\mbox{.}(2017)]%
        {DBLP:conf/nips/VaswaniSPUJGKP17}
\bibfield{author}{\bibinfo{person}{Ashish Vaswani}, \bibinfo{person}{Noam Shazeer}, \bibinfo{person}{Niki Parmar}, \bibinfo{person}{Jakob Uszkoreit}, \bibinfo{person}{Llion Jones}, \bibinfo{person}{Aidan~N. Gomez}, \bibinfo{person}{Lukasz Kaiser}, {and} \bibinfo{person}{Illia Polosukhin}.} \bibinfo{year}{2017}\natexlab{}.
\newblock \showarticletitle{Attention is All you Need}. In \bibinfo{booktitle}{\emph{Advances in Neural Information Processing Systems 30: Annual Conference on Neural Information Processing Systems 2017, December 4-9, 2017, Long Beach, CA, {USA}}}, \bibfield{editor}{\bibinfo{person}{Isabelle Guyon}, \bibinfo{person}{Ulrike von Luxburg}, \bibinfo{person}{Samy Bengio}, \bibinfo{person}{Hanna~M. Wallach}, \bibinfo{person}{Rob Fergus}, \bibinfo{person}{S.~V.~N. Vishwanathan}, {and} \bibinfo{person}{Roman Garnett}} (Eds.). \bibinfo{pages}{5998--6008}.
\newblock
\urldef\tempurl%
\url{https://proceedings.neurips.cc/paper/2017/hash/3f5ee243547dee91fbd053c1c4a845aa-Abstract.html}
\showURL{%
\tempurl}


\bibitem[Wang et~al\mbox{.}(2023)]%
        {DBLP:journals/tmm/WangWHLL23}
\bibfield{author}{\bibinfo{person}{Shuo Wang}, \bibinfo{person}{Zhihao Wu}, \bibinfo{person}{Xiaobo Hu}, \bibinfo{person}{Youfang Lin}, {and} \bibinfo{person}{Kai Lv}.} \bibinfo{year}{2023}\natexlab{}.
\newblock \showarticletitle{Skill-Based Hierarchical Reinforcement Learning for Target Visual Navigation}.
\newblock \bibinfo{journal}{\emph{{IEEE} Trans. Multim.}}  \bibinfo{volume}{25} (\bibinfo{year}{2023}), \bibinfo{pages}{8920--8932}.
\newblock
\href{https://doi.org/10.1109/TMM.2023.3243618}{doi:\nolinkurl{10.1109/TMM.2023.3243618}}


\bibitem[Wang et~al\mbox{.}(2024)]%
        {DBLP:conf/aaai/WangWHWLL24}
\bibfield{author}{\bibinfo{person}{Shuo Wang}, \bibinfo{person}{Zhihao Wu}, \bibinfo{person}{Xiaobo Hu}, \bibinfo{person}{Jinwen Wang}, \bibinfo{person}{Youfang Lin}, {and} \bibinfo{person}{Kai Lv}.} \bibinfo{year}{2024}\natexlab{}.
\newblock \showarticletitle{What Effects the Generalization in Visual Reinforcement Learning: Policy Consistency with Truncated Return Prediction}. In \bibinfo{booktitle}{\emph{Thirty-Eighth {AAAI} Conference on Artificial Intelligence, {AAAI} 2024, Thirty-Sixth Conference on Innovative Applications of Artificial Intelligence, {IAAI} 2024, Fourteenth Symposium on Educational Advances in Artificial Intelligence, {EAAI} 2014, February 20-27, 2024, Vancouver, Canada}}, \bibfield{editor}{\bibinfo{person}{Michael~J. Wooldridge}, \bibinfo{person}{Jennifer~G. Dy}, {and} \bibinfo{person}{Sriraam Natarajan}} (Eds.). \bibinfo{publisher}{{AAAI} Press}, \bibinfo{pages}{5590--5598}.
\newblock
\href{https://doi.org/10.1609/AAAI.V38I6.28369}{doi:\nolinkurl{10.1609/AAAI.V38I6.28369}}


\bibitem[Wu et~al\mbox{.}(2023)]%
        {DBLP:conf/nips/0001MDL23}
\bibfield{author}{\bibinfo{person}{Jialong Wu}, \bibinfo{person}{Haoyu Ma}, \bibinfo{person}{Chaoyi Deng}, {and} \bibinfo{person}{Mingsheng Long}.} \bibinfo{year}{2023}\natexlab{}.
\newblock \showarticletitle{Pre-training Contextualized World Models with In-the-wild Videos for Reinforcement Learning}. In \bibinfo{booktitle}{\emph{Advances in Neural Information Processing Systems 36: Annual Conference on Neural Information Processing Systems 2023, NeurIPS 2023, New Orleans, LA, USA, December 10 - 16, 2023}}, \bibfield{editor}{\bibinfo{person}{Alice Oh}, \bibinfo{person}{Tristan Naumann}, \bibinfo{person}{Amir Globerson}, \bibinfo{person}{Kate Saenko}, \bibinfo{person}{Moritz Hardt}, {and} \bibinfo{person}{Sergey Levine}} (Eds.).
\newblock
\urldef\tempurl%
\url{http://papers.nips.cc/paper\_files/paper/2023/hash/7ce1cbededb4b0d6202847ac1b484ee8-Abstract-Conference.html}
\showURL{%
\tempurl}


\bibitem[Yang et~al\mbox{.}(2019)]%
        {DBLP:conf/nips/YangDYCSL19}
\bibfield{author}{\bibinfo{person}{Zhilin Yang}, \bibinfo{person}{Zihang Dai}, \bibinfo{person}{Yiming Yang}, \bibinfo{person}{Jaime~G. Carbonell}, \bibinfo{person}{Ruslan Salakhutdinov}, {and} \bibinfo{person}{Quoc~V. Le}.} \bibinfo{year}{2019}\natexlab{}.
\newblock \showarticletitle{XLNet: Generalized Autoregressive Pretraining for Language Understanding}. In \bibinfo{booktitle}{\emph{Advances in Neural Information Processing Systems 32: Annual Conference on Neural Information Processing Systems 2019, NeurIPS 2019, December 8-14, 2019, Vancouver, BC, Canada}}, \bibfield{editor}{\bibinfo{person}{Hanna~M. Wallach}, \bibinfo{person}{Hugo Larochelle}, \bibinfo{person}{Alina Beygelzimer}, \bibinfo{person}{Florence d'Alch{\'{e}}{-}Buc}, \bibinfo{person}{Emily~B. Fox}, {and} \bibinfo{person}{Roman Garnett}} (Eds.). \bibinfo{pages}{5754--5764}.
\newblock
\urldef\tempurl%
\url{https://proceedings.neurips.cc/paper/2019/hash/dc6a7e655d7e5840e66733e9ee67cc69-Abstract.html}
\showURL{%
\tempurl}


\bibitem[Ye et~al\mbox{.}(2021)]%
        {DBLP:conf/nips/YeLKAG21}
\bibfield{author}{\bibinfo{person}{Weirui Ye}, \bibinfo{person}{Shaohuai Liu}, \bibinfo{person}{Thanard Kurutach}, \bibinfo{person}{Pieter Abbeel}, {and} \bibinfo{person}{Yang Gao}.} \bibinfo{year}{2021}\natexlab{}.
\newblock \showarticletitle{Mastering Atari Games with Limited Data}. In \bibinfo{booktitle}{\emph{Advances in Neural Information Processing Systems 34: Annual Conference on Neural Information Processing Systems 2021, NeurIPS 2021, December 6-14, 2021, virtual}}, \bibfield{editor}{\bibinfo{person}{Marc'Aurelio Ranzato}, \bibinfo{person}{Alina Beygelzimer}, \bibinfo{person}{Yann~N. Dauphin}, \bibinfo{person}{Percy Liang}, {and} \bibinfo{person}{Jennifer~Wortman Vaughan}} (Eds.). \bibinfo{pages}{25476--25488}.
\newblock
\urldef\tempurl%
\url{https://proceedings.neurips.cc/paper/2021/hash/d5eca8dc3820cad9fe56a3bafda65ca1-Abstract.html}
\showURL{%
\tempurl}


\bibitem[Ye et~al\mbox{.}(2023)]%
        {DBLP:conf/iclr/YeZAG23}
\bibfield{author}{\bibinfo{person}{Weirui Ye}, \bibinfo{person}{Yunsheng Zhang}, \bibinfo{person}{Pieter Abbeel}, {and} \bibinfo{person}{Yang Gao}.} \bibinfo{year}{2023}\natexlab{}.
\newblock \showarticletitle{Become a Proficient Player with Limited Data through Watching Pure Videos}. In \bibinfo{booktitle}{\emph{The Eleventh International Conference on Learning Representations, {ICLR} 2023, Kigali, Rwanda, May 1-5, 2023}}. \bibinfo{publisher}{OpenReview.net}.
\newblock
\urldef\tempurl%
\url{https://openreview.net/forum?id=Sy-o2N0hF4f}
\showURL{%
\tempurl}


\bibitem[Yu et~al\mbox{.}(2019)]%
        {DBLP:conf/corl/YuQHJHFL19}
\bibfield{author}{\bibinfo{person}{Tianhe Yu}, \bibinfo{person}{Deirdre Quillen}, \bibinfo{person}{Zhanpeng He}, \bibinfo{person}{Ryan Julian}, \bibinfo{person}{Karol Hausman}, \bibinfo{person}{Chelsea Finn}, {and} \bibinfo{person}{Sergey Levine}.} \bibinfo{year}{2019}\natexlab{}.
\newblock \showarticletitle{Meta-World: {A} Benchmark and Evaluation for Multi-Task and Meta Reinforcement Learning}. In \bibinfo{booktitle}{\emph{3rd Annual Conference on Robot Learning, CoRL 2019, Osaka, Japan, October 30 - November 1, 2019, Proceedings}} \emph{(\bibinfo{series}{Proceedings of Machine Learning Research}, Vol.~\bibinfo{volume}{100})}, \bibfield{editor}{\bibinfo{person}{Leslie~Pack Kaelbling}, \bibinfo{person}{Danica Kragic}, {and} \bibinfo{person}{Komei Sugiura}} (Eds.). \bibinfo{publisher}{{PMLR}}, \bibinfo{pages}{1094--1100}.
\newblock
\urldef\tempurl%
\url{http://proceedings.mlr.press/v100/yu20a.html}
\showURL{%
\tempurl}


\bibitem[Yuan et~al\mbox{.}(2022)]%
        {DBLP:conf/nips/YuanXYWWGX22}
\bibfield{author}{\bibinfo{person}{Zhecheng Yuan}, \bibinfo{person}{Zhengrong Xue}, \bibinfo{person}{Bo Yuan}, \bibinfo{person}{Xueqian Wang}, \bibinfo{person}{Yi Wu}, \bibinfo{person}{Yang Gao}, {and} \bibinfo{person}{Huazhe Xu}.} \bibinfo{year}{2022}\natexlab{}.
\newblock \showarticletitle{Pre-Trained Image Encoder for Generalizable Visual Reinforcement Learning}. In \bibinfo{booktitle}{\emph{Advances in Neural Information Processing Systems 35: Annual Conference on Neural Information Processing Systems 2022, NeurIPS 2022, New Orleans, LA, USA, November 28 - December 9, 2022}}, \bibfield{editor}{\bibinfo{person}{Sanmi Koyejo}, \bibinfo{person}{S.~Mohamed}, \bibinfo{person}{A.~Agarwal}, \bibinfo{person}{Danielle Belgrave}, \bibinfo{person}{K.~Cho}, {and} \bibinfo{person}{A.~Oh}} (Eds.).
\newblock
\urldef\tempurl%
\url{http://papers.nips.cc/paper\_files/paper/2022/hash/548a482d4496ce109cddfbeae5defa7d-Abstract-Conference.html}
\showURL{%
\tempurl}


\bibitem[Zhang et~al\mbox{.}(2024)]%
        {DBLP:conf/eccv/ZhangKSC24}
\bibfield{author}{\bibinfo{person}{Lixuan Zhang}, \bibinfo{person}{Meina Kan}, \bibinfo{person}{Shiguang Shan}, {and} \bibinfo{person}{Xilin Chen}.} \bibinfo{year}{2024}\natexlab{}.
\newblock \showarticletitle{PreLAR: World Model Pre-training with Learnable Action Representation}. In \bibinfo{booktitle}{\emph{Computer Vision - {ECCV} 2024 - 18th European Conference, Milan, Italy, September 29-October 4, 2024, Proceedings, Part {XXIII}}} \emph{(\bibinfo{series}{Lecture Notes in Computer Science}, Vol.~\bibinfo{volume}{15081})}. \bibinfo{publisher}{Springer}, \bibinfo{pages}{185--201}.
\newblock


\bibitem[Zhou et~al\mbox{.}(2018)]%
        {DBLP:journals/pami/ZhouLKO018}
\bibfield{author}{\bibinfo{person}{Bolei Zhou}, \bibinfo{person}{{\`{A}}gata Lapedriza}, \bibinfo{person}{Aditya Khosla}, \bibinfo{person}{Aude Oliva}, {and} \bibinfo{person}{Antonio Torralba}.} \bibinfo{year}{2018}\natexlab{}.
\newblock \showarticletitle{Places: {A} 10 Million Image Database for Scene Recognition}.
\newblock \bibinfo{journal}{\emph{{IEEE} Trans. Pattern Anal. Mach. Intell.}} \bibinfo{volume}{40}, \bibinfo{number}{6} (\bibinfo{year}{2018}), \bibinfo{pages}{1452--1464}.
\newblock
\href{https://doi.org/10.1109/TPAMI.2017.2723009}{doi:\nolinkurl{10.1109/TPAMI.2017.2723009}}


\bibitem[Zhou et~al\mbox{.}(2023)]%
        {DBLP:conf/nips/ZhouLJL23}
\bibfield{author}{\bibinfo{person}{Bohan Zhou}, \bibinfo{person}{Ke Li}, \bibinfo{person}{Jiechuan Jiang}, {and} \bibinfo{person}{Zongqing Lu}.} \bibinfo{year}{2023}\natexlab{}.
\newblock \showarticletitle{Learning from Visual Observation via Offline Pretrained State-to-Go Transformer}. In \bibinfo{booktitle}{\emph{Advances in Neural Information Processing Systems 36: Annual Conference on Neural Information Processing Systems 2023, NeurIPS 2023, New Orleans, LA, USA, December 10 - 16, 2023}}, \bibfield{editor}{\bibinfo{person}{Alice Oh}, \bibinfo{person}{Tristan Naumann}, \bibinfo{person}{Amir Globerson}, \bibinfo{person}{Kate Saenko}, \bibinfo{person}{Moritz Hardt}, {and} \bibinfo{person}{Sergey Levine}} (Eds.).
\newblock
\urldef\tempurl%
\url{http://papers.nips.cc/paper\_files/paper/2023/hash/bb203e938836544655996d1bb94a0fd7-Abstract-Conference.html}
\showURL{%
\tempurl}


\end{thebibliography}










\end{document}